\documentclass[letterpaper]{article} % DO NOT CHANGE THIS
\usepackage{aaai2026}  % DO NOT CHANGE THIS
\usepackage{times}  % DO NOT CHANGE THIS
\usepackage{helvet}  % DO NOT CHANGE THIS
\usepackage{courier}  % DO NOT CHANGE THIS
\usepackage[hyphens]{url}  % DO NOT CHANGE THIS
\usepackage{graphicx} % DO NOT CHANGE THIS
\usepackage{natbib}  % DO NOT CHANGE THIS AND DO NOT ADD ANY OPTIONS TO IT
\usepackage{caption} % DO NOT CHANGE THIS AND DO NOT ADD ANY OPTIONS TO IT
\usepackage{algorithm}
\usepackage{algorithmic}

\usepackage{newfloat}
\usepackage{listings}
\DeclareCaptionStyle{ruled}{labelfont=normalfont,labelsep=colon,strut=off} % DO NOT CHANGE THIS
\floatstyle{ruled}
\newfloat{listing}{tb}{lst}{}
\floatname{listing}{Listing}
\usepackage{xspace}
\usepackage{array} % required for text wrapping in tables
\usepackage{multirow}
\usepackage{makecell}
\usepackage{booktabs}
\usepackage{subfigure}
\usepackage[most]{tcolorbox}

\newcommand{\algorithmName}[1]{{#1}\xspace}
\newcommand{\datasetName}[1]{{#1}\xspace}
\newcommand{\aiarxiv}[0]{\datasetName{AIArxiv}}
\newcommand{\bioasq}[0]{\datasetName{BioASQ}}
\newcommand{\clapnq}[0]{\datasetName{ClapNQ}}
\newcommand{\miniwiki}[0]{\datasetName{MiniWiki}}
\newcommand{\watsonxdataset}[0]{\datasetName{WatsonxQA}} % PREPRINT
\newcommand\ob[1]{}%\textcolor{blue}{[OB: #1]}}
\newcommand\pt[1]{}%\textcolor{orange}{[PT: #1]}}
\newcommand\oh[1]{}%\textcolor{olive}{[OE: #1]}}

\newcommand{\tableRef}[1]{Table~\ref{#1}}
\newcommand{\sectionRef}[1]{\S\ref{#1}}
\newcommand{\figureRef}[1]{Figure~\ref{#1}}

\newcommand{\supp}[0]
{Appendix\xspace}
\newcommand{\greedyModel}[0]{\algorithmName{Greedy-M}}
\newcommand{\greedyRetriever}[0]{\algorithmName{Greedy-R}}
\newcommand{\greedyRetrieverCC}[0]{\algorithmName{Greedy-R-CC}}

\newcommand{\tpe}[0]{\algorithmName{TPE}}
\newcommand{\random}[0]{\algorithmName{Random}}
\newcommand{\greedym}[0]{\algorithmName{Greedy-M}}
\newcommand{\greedyr}[0]{\algorithmName{Greedy-R}}
\newcommand{\greedyrcc}[0]{\algorithmName{Greedy-R-CC}}

\newcommand{\chunksize}[0]{Chunk Size\xspace}
\newcommand{\chunkoverlap}[0]{Chunk Overlap\xspace}
\newcommand{\embmodel}[0]{Embedding Model}

\newcommand{\topk}[0]{Top-K\xspace}
\newcommand{\genmodel}[0]{Generative Model}

\newcommand{\metricName}[1]{{#1}\xspace}
\newcommand{\llmaaj}[0]{\metricName{LLMaaJ-AC}}
\newcommand{\lexical}[0]{\metricName{Lexical-AC}}
\newcommand{\faithfulness}[0]{\metricName{Lexical-FF}}

\newcommand{\modelName}[1]{{#1}\xspace}
\newcommand{\lamma}[0]{\modelName{Llama}}
\newcommand{\granite}[0]{\modelName{Granite}}
\newcommand{\mistral}[0]{\modelName{Mistral}}

\title{An Analysis of Hyper-Parameter Optimization Methods\\for Retrieval Augmented Generation}

\author{    
    Matan Orbach, 
    Ohad Eytan,
    Benjamin Sznajder,
    Ariel Gera, 
    Odellia Boni,
    \\
    Yoav Kantor, 
    Gal Bloch,
    Omri Levy,
    Hadas Abraham, 
    Nitzan Barzilay,
    \\
    Eyal Shnarch, 
    Michael E. Factor,
    Shila Ofek-Koifman,
    Paula Ta-Shma, 
    Assaf Toledo
}

\affiliations{
    IBM Research
    matano@il.ibm.com
}

\begin{document}

\maketitle

\begin{abstract}
Optimizing Retrieval-Augmented Generation (RAG) configurations for specific tasks is a complex and resource-intensive challenge.
Motivated by this challenge, frameworks for RAG hyper-parameter optimization (HPO) have recently emerged, yet their effectiveness has not been rigorously benchmarked. 
To fill this gap, we present a comprehensive study involving five HPO algorithms over five datasets from diverse domains, including a newly curated real-world  product documentation dataset.
Our study explores the largest RAG HPO search space to date that includes full grid-search evaluations, and uses three evaluation metrics as optimization targets.
Analysis of the results shows that RAG HPO can be done efficiently, either greedily or with random search, and that it significantly boosts RAG performance for all datasets. For greedy HPO approaches, we show that optimizing model selection first is preferable to the common practice of following the RAG pipeline order during optimization.
\end{abstract}

%\begin{links}
%    \link{Code}{https://aaai.org/example/code}
%    \link{Datasets}{https://huggingface.co/datasets/ibm-research/watsonxDocsQA}
%    \link{Extended version}{https://aaai.org/example/extended-version}
%\end{links}

\section{Introduction}
In the \textbf{Retrieval-Augmented Generation (RAG)} paradigm, a generative LLM answers user questions using a retrieval system which provides relevant context from a corpus of documents \citep{rag1, rag3, rag2, rag4}.
By relying on a dedicated retrieval component, RAG solutions focus LLMs on grounded data, reducing the likelihood of dependence on irrelevant preexisting knowledge.

The popularity of RAG is largely thanks to its modular design, allowing full control over which data sources to pull data from and how to process that data.
While advantageous, this modularity also means that practitioners are faced with a wide array of decisions when designing their RAG pipelines.
One such choice is which generative LLM to use;
other choices pertain to parameters of the retrieval system, such as how many items to retrieve per input question, how to rank them, and so forth.

Furthermore, evaluating even a single RAG configuration is costly in terms of time and funds: the embedding step as part of corpus indexing is compute-intensive; generating answers using LLMs is also a demanding task, especially for large benchmarks; and evaluation with LLM-as-a-Judge (LLMaaJ) adds another costly round of inference.
As a result, exhaustively exploring the exponential search space of RAG parameters is prohibitively expensive. 
At the same time, suboptimal parameter choices may significantly harm result quality. 

A promising approach to this challenge is \textbf{hyper-parameter optimization (HPO)} for RAG \citep{fu-etal-2024-autorag, kim2024autorag}, 
which aims to identify high-performing configurations by systematically evaluating a  subset of the search space.
Existing methods range from established HPO algorithms to simple random sampling.
Importantly, despite growing interest, the effectiveness of HPO in realistic RAG scenarios has not been rigorously tested.

\begin{figure}[t]
  \includegraphics[width=1\columnwidth]{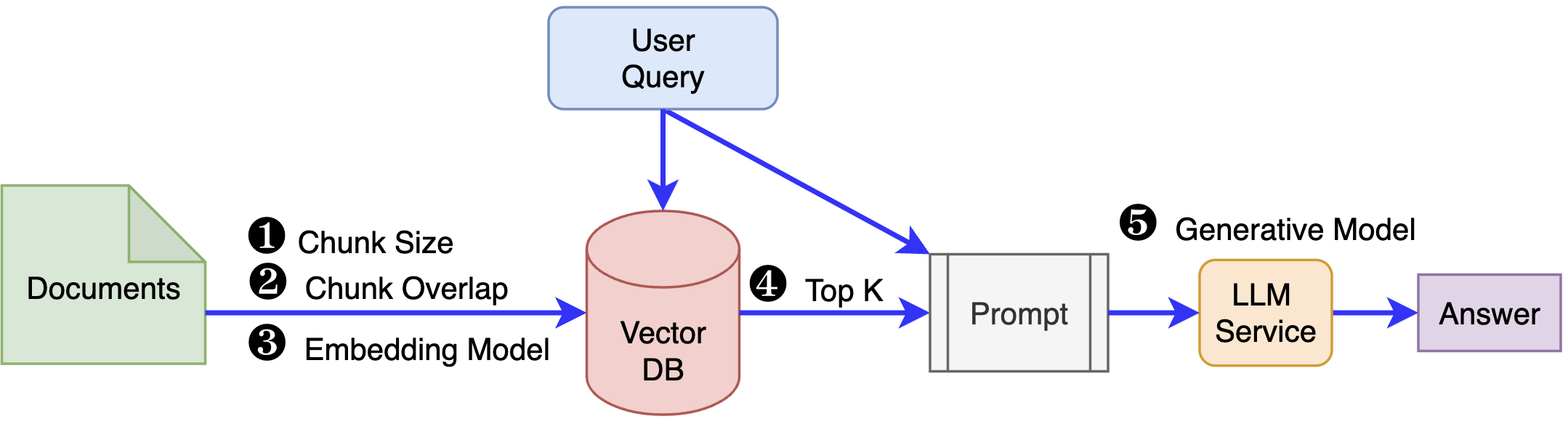}
  \caption {We study hyper parameter optimization over a RAG pipeline with $5$ parameters. The explored search space includes $162$ RAG configurations formed from combinations of the depicted hyper parameters.}
  \label{fig:flow}
\end{figure}

Our work addresses this gap through a comprehensive study of HPO for RAG in a setup that mirrors real-world usage, where a dataset is provided up front for experimentation, and unseen queries arrive after deployment.
The best RAG configuration is therefore selected by development set performance and evaluated on held out test data. 
To the best of our knowledge, this is the first study that considers RAG HPO in such a setup.

The scope of our experiments includes multiple datasets, evaluation metrics and algorithms.
To represent diverse use cases, we evaluate across several domains: scientific \citep{eibich2024aragog}, biomedical \citep{krithara2023bioasq}, Wikipedia \citep{rosenthal2024clapnq, smith2008question} and a newly curated enterprise product documentation dataset that we open-source as part of this work.\footnote{\url{http://huggingface.co/datasets/ibm-research/watsonxDocsQA}} %  ANONYMIZE
Additionally, recognizing that different applications prioritize distinct performance metrics -- such as answer correctness or faithfulness -- our study evaluates multiple RAG optimization objectives, implemented through two alternative approaches: lexical overlap metrics and LLMaaJ.

Our evaluation compares five HPO algorithms: Tree-Structured Parzen Estimators (TPE) \citep{tpe}, three greedy variants, and random search. 
Also included are baseline grid search results for both the development and test sets, over the full search space.\footnote{The grid results are at \url{https://github.com/IBM/rag-hpo-bench}.}
These are crucial for establishing upper bounds on HPO performance for RAG, and understanding its generalization capability. 

While the inclusion of grid search enables a comparison to the best possible result,  it poses computational constraints on the size of the explored search space. 
Nonetheless, our search space is the largest considered for RAG HPO to date while still including a comparison to full grid search. 

The search space comprises of $162$ RAG configurations derived from five core RAG parameters (see \figureRef{fig:flow}): chunk size and overlap, which control how documents are split, the embedding model used to encode chunks in a vector database, the number of retrieved chunks included as context when answering a question, and the generative model that produces the answer. 
Exploring an expanded search space introduced by more complex RAG pipelines, such as agentic workflows, poses significant computational challenges for exhaustive grid search. Consequently, this aspect is deferred to future work. 

Our evaluation addresses multiple aspects of HPO for RAG, including the convergence properties of the algorithms, the impact of the objective metric on the best configuration, and an analysis of HPO overall cost. We also explore reducing that cost through development set sampling.

% ADD something on the dev set sampling and theincluded analysis

In summary, our main contributions are as follows:
(i) comprehensive benchmarking of RAG HPO in a realistic generalization setup, over the largest search space with full grid-search evaluations to-date, showing that RAG HPO can be done efficiently, either greedily or with simple random search, and that it significantly boosts RAG performance for all datasets; 
(ii) a detailed analysis of the results, exploring the connections between the optimized parameters, the dataset and the optimization objective. For greedy HPO approaches, we show that the order in which the parameters are optimized is of great importance;
(iii) new open-source resources: the full grid search results of our experiments, and an enterprise product documentation RAG dataset.

\section{Related Work}
\label{sec:related_work}

Within the open-source community, several tools offer out-of-the-box HPO algorithms for RAG. 
AutoRAG \citep{kim2024autorag} adopts a greedy approach, optimizing one RAG parameter at a time following the sequential pipeline order.\footnote{\url{https://github.com/Marker-Inc-Korea/AutoRAG}} 
RAGBuilder employs TPE for HPO.\footnote{\url{https://github.com/KruxAI/ragbuilder}}
Additionaly, RAG-centric libraries such as LlamaIndex\footnote{\url{https://www.llamaindex.ai/}}  support HPO by integrating general-purpose optimization frameworks like RayTune \cite{liaw2018tune}, optuna \cite{optuna} and hyperopt \cite{hyperopt}.

Other works investigate the impact of RAG hyper-parameters without automated optimization.  
For example, \citet{crud-rag} and \citet{wang2024searching} focus on manual tuning of RAG hyper-parameters, while \citet{ragbench} evaluates multiple configurations via grid search.
Their studies highlight the critical role of hyper-parameter tuning in RAG and motivate the need for automated RAG HPO.

Another interesting line of work \citep{fu-etal-2024-autorag} is motivated by a setting of online feedback from users. It describes an online HPO algorithm that iteratively updates the reward for the various RAG parameters based on small batches of queries. In contrast, our work addresses offline HPO, where optimization is performed on full benchmark datasets prior to best configuration deployment.

More recently, \citet{barker2025multi_objective_optimization} introduced multi-objective HPO for RAG. The studied methods select a set of RAG configuration deemed Pareto-optimal by multiple metrics. Choosing the best configuration from this set remains an open challenge. Our study instead focuses on single-objective HPO returning a single configuration as output.

Despite these efforts, still missing is a systematic evaluation of HPO algorithms across diverse datasets under realistic conditions where optimized configurations are tested on held-out sets.
This gap is the primary focus of our work. 
Building upon existing approaches, our evaluation prioritizes HPO algorithms already considered in the context of RAG, over introducing alternatives such as BOHB \cite{falkner2018bohb} or SMAC \citep{lindauer2022smac3}.
Specifically, the greedy algorithms we explore 
%in \sectionRef{sec:hpo} 
resemble those used by \citet{kim2024autorag}, and the \tpe algorithm we assess is the same one used by RAGBuilder. 

\section{Experimental Setup}

%The architecture of different RAG pipelines is diverse, with some components common to most architectures. 
%We focus on these common parts, which we represent by the following linear flow.
\subsection{Search space}
\label{sec:search_space}

\begin{table*}
   \centering
   \begin{tabular}{>{\raggedright\arraybackslash}p{0.27\linewidth}>{\raggedright\arraybackslash}p{0.12\linewidth}p{8cm}}
     \hline
     \textbf{Hyper Parameter} & 
 \textbf{RAG step}&
    \textbf{Values}\\ 
     \hline
     \chunksize{} (\# Tokens) & Chunking  &256, 384, 512\\
     \hline
     \chunkoverlap{} (\% Tokens) &  Chunking& 0\%, 25\%\\
     \hline
     \multirow[t]{3}{*}{\embmodel} & 
     Embedding &  
     \makecell[tl]{\textit{multilingual-e5-large}  \cite{multilingual-e5-large} \\ \textit{bge-large-en-v1.5} \cite{bge_embedding} \\ \textit{granite-embedding-125M-english} \cite{ibm2024graniteembedding125m}}
    \\
     \hline
     \topk{} (\# Chunks to retrieve) & Retrieval & 3, 5, 10\\ 
     
     \hline
     \multirow[t]{3}{*}{\genmodel}& 
     Generation& \makecell[tl]{\textit{Llama-3.1-8B-Instruct} \cite{meta2024llama3.1} \\ \textit{Mistral-Nemo-Instruct-2407} \cite{mistral2024nemo}\\  \textit{Granite-3.1-8B-instruct} \cite{granite2024}} \\ \hline
   \end{tabular}
   \caption{\label{tab:hyper-param}
     The hyper parameters explored in our search space, and their values.
   }
   \end{table*}

In our explored RAG pipeline (see \figureRef{fig:flow}), processing starts with \textbf{chunking} the input documents into smaller chunks, based on two parameters: the \textit{size} of each chunk in tokens, and the \textit{overlap} between consecutive chunks.
This is followed by representing each chunk with a dense vector created by an \textbf{embedding} model -- our third parameter.
The vectors are stored in a vector-database,\footnote{Milvus \citep{milvus} with default settings; index type is HNSW \cite{hnsw}.} alongside their original text. 
Upon receiving a query, the top $k$ relevant chunks are retrieved (\textbf{retrieval}); $k$ being our fourth parameter. Lastly, a prompt containing the query and the retrieved chunks is passed to a \textit{generative model} (our fifth parameter) to create an answer (\textbf{generation}). Greedy decoding was used throughout all experiments. The prompts were fixed to RAG prompts tailored to each model.\footnote{See Appendix \ref{sec:app_generation_details}.} 

%The flow of our RAG pipeline together with the hyper parameters we used (see below) is depicted in Fig~\ref{fig:flow}.

The specific values considered for each of the five hyper-parameters are described in Table~\ref{tab:hyper-param}. In total, the search space has $3*2*3*3*3 =162$ possible {\bf RAG configurations}. Dividing into stages, we get $18$ ($3*2*3$) different configurations for data indexing (chunking and embedding) and $9$ ($3*3$) different configurations for answering (retrieval and generation).
The values chosen for the chunk size, overlap, and top-k reflect common practices (see Appendix \ref{sec:search_space_selection} for details). 
Popular open source models were selected as options for the embedding and generative models. 
We opted to focus on LLMs that are similar in size, since otherwise an obvious strategy is to simply discard the smaller less capable models from the search. 
% Initially, we considered combining LLMs of different sizes in the search space. However, this may result in a less informative HPO setup, where the obvious strategy is to immediately discard the smaller and less capable models from the search space. Thus, we opted to focus on LLMs that are similar in size. As above, given the prohibitive costs of grid search, we opted for a set of moderately sized models: Llama-3.1-8B-Instruct, Mistral-Nemo-Instruct-2407 (12B parameters) and Granite-3.1-8B-instruct. 

Our experiments involve performing a full grid search -- i.e., evaluating all possible configurations -- in order to establish upper bound baselines for the optimization strategies. Hence, due to computational constraints we avoid using an even larger search space, and choose a set of moderately sized LLMs as our generators.
Still, to the best of our knowledge, our search space is the largest considered to date that compares to full grid-search.

\subsection{Datasets}
\label{sec:datasets}

\begin{table}
  \centering
  \begin{tabular}
  {
  >{\raggedright\arraybackslash}p{0.22\columnwidth}
  >{\raggedright\arraybackslash}p{0.2\columnwidth}
  >{\centering\arraybackslash}p{0.12\columnwidth}
  >{\centering\arraybackslash}p{0.1\columnwidth}
  >{\centering\arraybackslash}p{0.1\columnwidth}
  }
    \toprule
    \textbf{Dataset}& \textbf{Domain} & \textbf{\#Doc}&\textbf{\#Dev}& \textbf{\#Test} \\
    \midrule
    \aiarxiv& Papers &$2673$   &$41$ & $30$\\
    \bioasq & Biomedical &$40181$& $1000$& $150$\\
    \miniwiki& Wikipedia &$3200$& $663$& $150$\\
    \clapnq & Wikipedia &$178890$& $1000$ & $150$\\
    \watsonxdataset{} & Business &$5534$& $45$ & $30$\\
    \bottomrule
  \end{tabular}
  \caption{\label{alldataset}
    Properties of the RAG Datasets in our experiments: the number of documents within the corpus (\textbf{\#Doc}), and counts of QA pairs in the \textbf{\#Dev} and \textbf{\#Test} sets.
  }
\end{table}

Each RAG dataset is comprised of a corpus of documents and a benchmark of QA pairs, with most also annotating the document(s) with the correct answer.
Below are the RAG datasets we used:

\paragraph{\aiarxiv} This dataset was derived from the ARAGOG benchmark \citep{eibich2024aragog} of technical QA pairs over a corpus of machine learning papers from ArXiv.\footnote{\url{https://huggingface.co/datasets/jamescalam/ai-arxiv2}}
As gold documents are not annotated in ARAGOG dataset, we added such labels where they could be found, obtaining $71$ answerable QA pairs out of $107$ in the original benchmark.

% TODO Final version - update the description of BioASQ, the split follows the split in HF.
\paragraph{\bioasq} \citep{krithara2023bioasq}
A subset of the \bioasq Challenge train set.\footnote{\url{https://huggingface.co/datasets/rag-datasets/rag-mini-bioasq}}
Its corpus contains $40200$ passages extracted from clinical case reports. 
The corresponding benchmark of $4200$ QA pairs includes multiple gold documents per question. 

\paragraph{\miniwiki} 
A benchmark of $918$ QA pairs over Wikipedia derived from \citet{smith2008question}.\footnote{\url{https://huggingface.co/datasets/rag-datasets/rag-mini-wikipedia}} The contents are mostly factoid questions with short answers. This dataset has no gold document labels. 

\paragraph{\clapnq} \cite{rosenthal2024clapnq} A subset of the Natural Questions (NQ) dataset \citep{kwiatkowski2019natural} on Wikipedia pages, of questions that have long answers. 
The original benchmark contains both answerable and unanswerable questions. For our analysis we consider only the former. \clapnq{} dataset consists of $178890$ passage texts generated from $4293$ pages. These passages constitute the input to the pipeline.

% ANONYMIZE
\paragraph{\watsonxdataset} (ProductDocs) 
%A contribution of this work, 
A new open-source dataset and benchmark based on enterprise product documentation, consisting of $5534$ passage texts created from $1144$ HTML product documentation pages.\footnote{\url{http://huggingface.co/datasets/ibm-research/watsonxDocsQA}}
These passages serve as the RAG pipeline input. 
The benchmark includes $75$ QA pairs and gold document labels, of which $25$ were generated by two subject matter experts, and the rest were synthetically produced and then manually filtered. All QA pairs were additionally reviewed by two of the authors, ensuring high data quality.
Further details are in Appendix~\ref{app:watsonx_details}.
%\footnote{More details in .} 

Overall, these datasets exhibit variability in many aspects. They represent diverse domains -- research papers, biomedical documents, wikipedia pages and enterprise data (see question examples in \tableRef{tab:dataset_examples}). They also vary in question and answer lengths; for example, \miniwiki has relatively short answers, while \clapnq was purposely built with long gold answers. Corpus sizes also vary, representing real-world retrieval scenarios over small or large sets of documents.

Every benchmark was split into development and test sets. 
To keep computations tractable, the number of questions in the large benchmarks (\bioasq and \clapnq) was limited to $1000$ for development and $150$ for test. Table~\ref{alldataset} lists the corpora benchmark sizes and domains. 

\begin{table}[t]
  \centering
  \begin{tabular}{
  p{0.20\linewidth}p{0.70\linewidth}
  }
    \toprule
    \textbf{Dataset}& \textbf{Example Question}\\
    \midrule
    \aiarxiv&
    What significant improvements does BERT bring to the SQuAD v1.1,v2.0 and v13.5 tasks compared to prior models?\\
    \bioasq      & 
    What is the implication of histone lysine methylation in medulloblastoma?\\
    \miniwiki &
    Was Abraham Lincoln the sixteenth President of the United States?\\
    \clapnq &
    Who is given credit for inventing the printing press?\\
    \watsonxdataset{}      &
    What tuning parameters are available for IBM foundation models?\\
    \bottomrule
  \end{tabular}
  \caption{\label{app:dataset_examples}
    One question example from each  dataset.
  }
  \label{tab:dataset_examples}
\end{table}

% Question examples are in Appendix \ref{sec:appendix-datasets}.

\subsection{Metrics}
The following metrics were used in our experiments:
\label{sec:evaluation}
{\bf Retrieval quality} was measured using {\bf context correctness} with Mean Reciprocal Rank
\cite{voorhees-tice-2000-trec}.\footnote{
Note that as labeling is at the document level, any chunk from the gold document is considered as a correct prediction, even if it does not include the answer to the question. 
}
{\bf Answer faithfulness} (\emph\faithfulness) measures whether a generated answer remained faithful to the retrieved contexts, with lexical token precision. 
{\bf Answer correctness} compares generated and gold answers, assessing {\bf overall pipeline quality}, and is measured in two ways:
First, a fast, lexical, implementation based on token recall (\emph{\lexical}), which provides a good balance between speed and quality \cite{adlakha-etal-2024-evaluating}.
 Second, a standard LLM-as-a-Judge answer correctness (\emph{\llmaaj}) implementation from the RAGAS library \cite{ragas}, with
 GPT4o-mini \cite{openai2024gpt4omini} as its backbone. 
 This implementation performs $3$ calls to the LLM on every invocation, making it much slower and more expensive than the lexical variant.

Given a benchmark, all metrics were computed per-question. Averaging the per-question scores yields the overall metric score.

We note that all benchmarked HPO approaches are metric-agnostic - they are not tailored to any specific metric, nor a specific performance axis (such as answer correctness). Similarly, HPO can be applied to multiple RAG metrics at once, forming a single optimization objective.%, and optimized with the considered methods. 
%MO While iterating over a benchmark, context correctness is computed per question after retrieval, and answer correctness is computed after answer generation. Averaging these scores over all benchmark instances determines the global scores of the RAG configuration on the dataset.

\subsection{HPO Algorithms}
\label{sec:hpo}

An HPO algorithm takes as input a RAG search space, a dataset (corpus and benchmark), and one evaluation metric designated as the optimization objective.
Its goal is to find the RAG configuration that achieves the highest performance on the dataset with respect to the objective.

The HPO algorithms in our experiments operate iteratively. At each iteration, the algorithm reviews the scores of all previously explored configurations and selects an unexplored configuration to evaluate next. 
To simulate a constrained exploration budget, the algorithm terminates after a fixed number of iterations and returns the best performing configuration.
An efficient HPO algorithm identifies a top-performing configuration with minimal iterations.

We examine two categories of HPO algorithms: (i) standard HPO algorithms that are not specifically tailored to RAG; (ii) RAG-aware greedy algorithms that leverage some knowledge of the components within the optimized RAG pipeline. 
All algorithms optimize answer correctness unless explicitly noted otherwise.

\paragraph{Standard algorithms} Our first standard HPO algorithm is \emph{\tpe}, using an implementation from hyperopt \cite{hyperopt}, with five random initialization iterations, and otherwise the default settings. 
The second algorithm (\emph{\random}) disregards results from prior iterations and uniformly selects an unexplored RAG configuration.

\paragraph{RAG-Aware greedy algorithms} 
The second category of algorithms assumes an ordered list of search space parameters ranked by their presumed impact on RAG performance.
These algorithms take a greedy approach: they iterate through the parameters list, optimizing one parameter at a time, assuming that optimizing high impact parameters first accelerates convergence to a strong configuration. 
When optimizing a parameter $p$, the algorithm uses fixed values  for all preceding parameters and evaluates  all possible values pf $p$,  with random values assigned to all following parameters.
The value of $p$ yielding the best objective score is picked, and the algorithm continues to the next parameter.

The greedy algorithms differ solely in their parameter ordering.
\textbf{Model-first ordering} (\emph{\greedym}) optimizes the generative and embedding models first, assuming they are more important: \genmodel, \embmodel, \chunksize, \chunkoverlap, \topk.
\textbf{Retrieval-first} (\emph{\greedyr}) is a prevalent option following the RAG pipeline structure, starting with retrieval optimization (still with model first), then generation: \embmodel, \chunksize, \chunkoverlap, \genmodel{}  and \topk.
\textbf{Retrieval-first with context correctness} (\emph{\greedyrcc}) uses the same order, yet optimizes the  retrieval-related parameters with a context correctness metric evaluated solely on the retrieval results, saving the costs of LLM inference until all retrieval parameters are chosen. Remaining parameters are then optimized with answer correctness.

\subsection{Setup}

Our experimental design reflects a realistic use case: an HPO algorithm is executed over a benchmark (a development set) and the best RAG configuration is chosen for deployment; the deployed configuration is expected to generalize to unseen questions -- simulated here via a test set.
Specifically, each HPO algorithm ran on each development set for $10$ iterations.
After every iteration, the best configuration on the development set was evaluated on the test set, enabling  a per-iteration generalization analysis.
Since all algorithms involve a random element, each run was repeated with $10$ different random seeds.

\section{Results}
\label{sec:results}

\begin{table}
  \centering
\begin{tabular}
{
>{\raggedright\arraybackslash}p{0.17\columnwidth}
>{\centering\arraybackslash}p{0.08\columnwidth}
>{\centering\arraybackslash}p{0.08\columnwidth}
>{\centering\arraybackslash}p{0.08\columnwidth}
>{\centering\arraybackslash}p{0.08\columnwidth}
>{\centering\arraybackslash}p{0.08\columnwidth}
>{\centering\arraybackslash}p{0.08\columnwidth}
}
\toprule
& \multicolumn{3}{c}{\textbf{\llmaaj}} 
& \multicolumn{3}{c}{\textbf{\lexical}} 
\\
\cmidrule(lr){2-7}
\textbf{Dataset} 
& \textbf{Worst} 
& \textbf{Best} 
& \textbf{SE} 
& \textbf{Worst} 
& \textbf{Best}
& \textbf{SE}\\
\midrule
%\aiarxiv        & 0.36 & 0.62 & 0.028 & 0.40 & 0.66 & 0.041 \\
%\bioasq         & 0.43 & 0.56 & 0.008 & 0.49 & 0.63 & 0.008 \\
%\miniwiki       & 0.32 & 0.51 & 0.018 & 0.61 & 0.85 & 0.01 \\
%\clapnq         & 0.46 & 0.57 & 0.008 & 0.34 & 0.61 & 0.008 \\
%\watsonxdataset & 0.52 & 0.76 & 0.033 & 0.74 & 0.87 & 0.038\\
\aiarxiv        & 0.36 & 0.62 & 0.03 & 0.40 & 0.66 & 0.04 \\
\bioasq         & 0.43 & 0.56 & 0.01 & 0.49 & 0.63 & 0.01 \\
\miniwiki       & 0.32 & 0.51 & 0.02 & 0.61 & 0.85 & 0.01 \\
\clapnq         & 0.46 & 0.57 & 0.01 & 0.34 & 0.61 & 0.01 \\
\watsonxdataset & 0.52 & 0.76 & 0.03 & 0.74 & 0.87 & 0.04 \\
\bottomrule
  \end{tabular}
  \caption{\label{confidenceinterval}
  \textbf{Worst} and \textbf{Best} configuration scores per dataset on the development set, reported for both \llmaaj and \lexical metrics. Also shown is the maximum standard error (\textbf{SE}) observed across all configurations.
  }
\end{table}

\begin{figure}[t]
  \includegraphics[width=1\columnwidth]{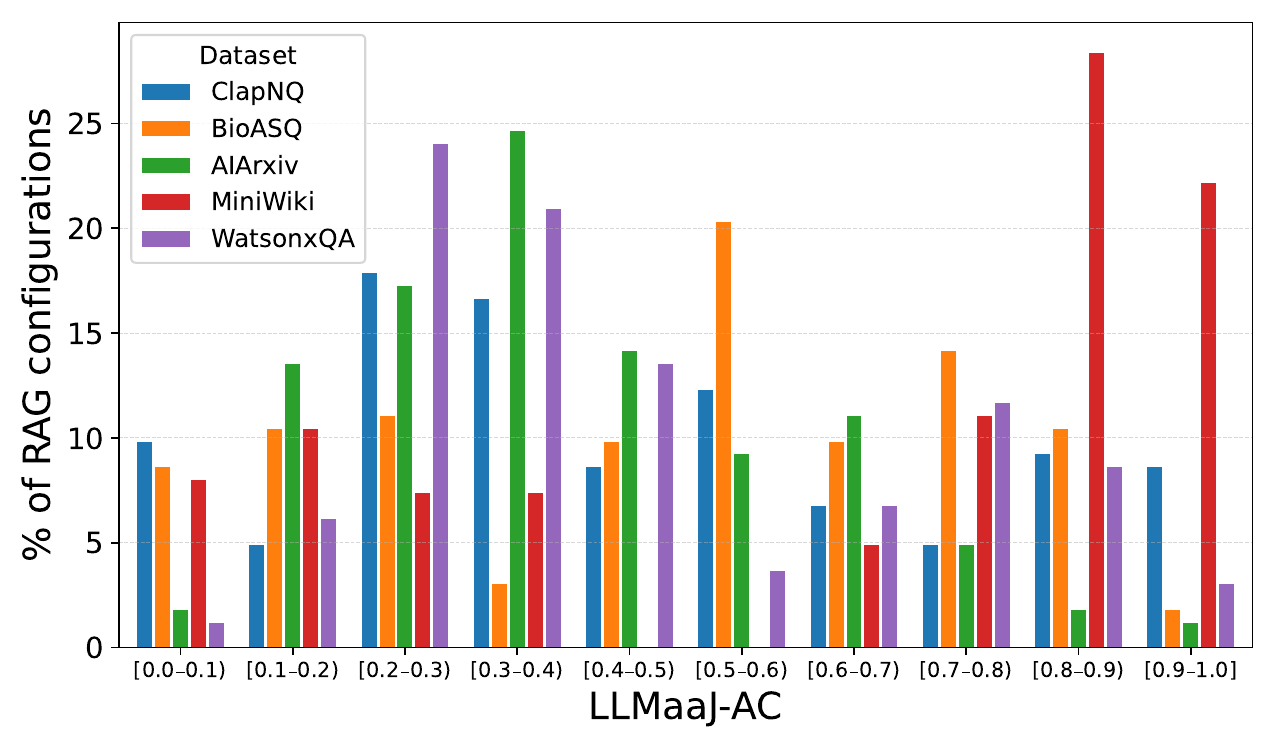}
  \caption {
  The distribution of configurations across bins for the normalized \llmaaj metric on the development sets. Most datasets have a few top-performing configurations.}
  \label{fig:configurations_distribution_ragas}
\end{figure}
\begin{figure*}[h!]
  \centering
  \subfigure[\llmaaj]{
  %\begin{minipage}{\columnwidth}
    \begin{tabular}{c}
        % trim=left bottom right top
            \includegraphics[clip, trim=0.18cm 27.5cm 1.2cm 3.5cm, width=0.95\columnwidth]{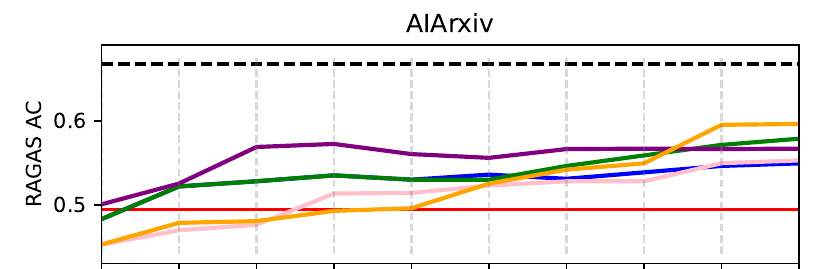}\\
            \includegraphics[clip, trim=0.18cm 21.5cm 1.2cm 9.5cm, width=0.95\columnwidth]{figures/test_results/test-results_data-full_AC-RAGAS.pdf}\\
            \includegraphics[clip, trim=0.18cm 15.5cm 1.2cm 15.5cm, width=0.95\columnwidth]{figures/test_results/test-results_data-full_AC-RAGAS.pdf}\\
            \includegraphics[clip, trim=0.18cm 9.6cm 1.2cm 21.5cm, width=0.95\columnwidth]{figures/test_results/test-results_data-full_AC-RAGAS.pdf}\\
            \includegraphics[clip, trim=0.18cm 2.5cm 1.2cm 27.3cm, width=0.95\columnwidth]{figures/test_results/test-results_data-full_AC-RAGAS.pdf}\\
        %\caption{LLMaaJ answer correctness}
        \end{tabular}
    \label{fig:ragas_full_data_test_results}
  %\end{minipage}%
  }
  \subfigure[\lexical]{
  %\begin{minipage}{\columnwidth}
        \begin{tabular}{c}
            \includegraphics[clip, trim=0.18cm 27.5cm 1.2cm 3.5cm, width=0.95\columnwidth]{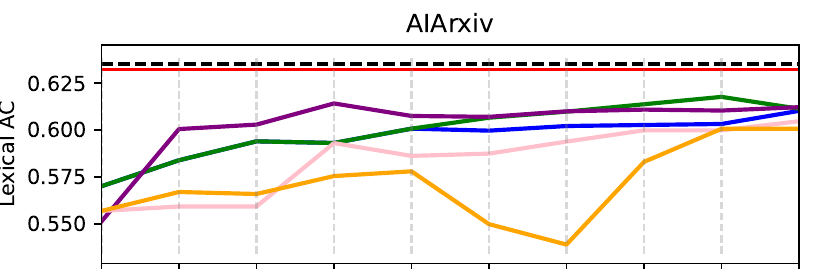}\\
            \includegraphics[clip, trim=0.18cm 21.5cm 1.2cm 9.5cm, width=0.95\columnwidth]{figures/test_results/test-results_data-full_AC-lexical.pdf}\\
            \includegraphics[clip, trim=0.18cm 15.5cm 1.2cm 15.5cm, width=0.95\columnwidth]{figures/test_results/test-results_data-full_AC-lexical.pdf}\\
            \includegraphics[clip, trim=0.18cm 9.6cm 1.2cm 21.5cm, width=0.95\columnwidth]{figures/test_results/test-results_data-full_AC-lexical.pdf}\\
            \includegraphics[clip, trim=0.18cm 2.5cm 1.2cm 27.3cm, width=0.95\columnwidth]{figures/test_results/test-results_data-full_AC-lexical.pdf}
        \end{tabular}

    %\caption{Lexical answer correctness}
    \label{fig:lexical_full_data_test_results}
 % \end{minipage}
 }
  \includegraphics[clip, trim=0cm 0cm 0cm 0cm, width=0.7\textwidth]{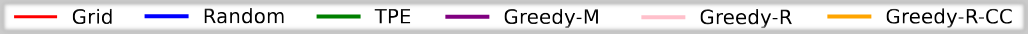}
  \caption{Per-iteration performance of all HPO algorithms on the test sets of five datasets, optimizing answer correctness computed with an LLMaaJ metric (a) and a lexical metric (b). The dashed black lines show the best achievable performance. The red lines are the performance of the best configuration chosen with development set evaluation, on the test set.}
  \label{fig:full_data_test_results}
\end{figure*}

%\begin{figure}[h!]

\subsection{Grid Search}
%\paragraph{Grid results} 
We conducted a comprehensive grid search over all $162$ configurations (including $18$ different indexes), across the development and test sets from all datasets. 
The worst and best performing configuration scores for each dataset, on the development set, are presented in Table~\ref{confidenceinterval}.\footnote{For results with \faithfulness see Appendix \ref{app:additional_results}.}
There is a substantial gap between the two extremes.

% 
% We examine the difference between worst and best performing configuration scores for each dataset with respect to $95\%$ confidence interval size of the score. As in all cases the confidence intervals of best and worst configuration scores do not overlap, we conclude that the difference in configurations performance is statistically significant.

The exhaustive grid search enables a deeper analysis of the configuration landscape, including the proportion of high and low performing configurations. 
%Having results over the entire search space allows computing the percentage of good or bad configurations present in the defined search space. 
To quantify this, we computed a min-max normalized score per dataset and metric, binned the scores uniformly, and assigned each configuration to a bin by its normalized metric score.  \figureRef{fig:configurations_distribution_ragas} shows the distribution of configurations across bins for the \llmaaj metric.
Notably, for most datasets, there are a few top-performing configurations. 
One example is the \bioasq dataset with fewer then  $5\%$ of configurations in the top two bins. 
In contrast, the \miniwiki dataset exhibits a dense cluster of good configurations, suggesting that a good configuration will be easy to find.
These trends exemplify that the difficulty of the HPO setup is dataset and metric dependent.\footnote{For results with \lexical and \faithfulness see Appendix \ref{app:additional_results}.}

%\footnote{Appendix \ref{app:additional_results} has results for the \lexical and \faithfulness.}

\begin{figure*}[h]
  \centering
  \subfigure[\llmaaj]{
    \centering
    % trim=left bottom right top
    \includegraphics[clip, trim=0.18cm 0.1cm 1.2cm 0cm, width=0.98\columnwidth]{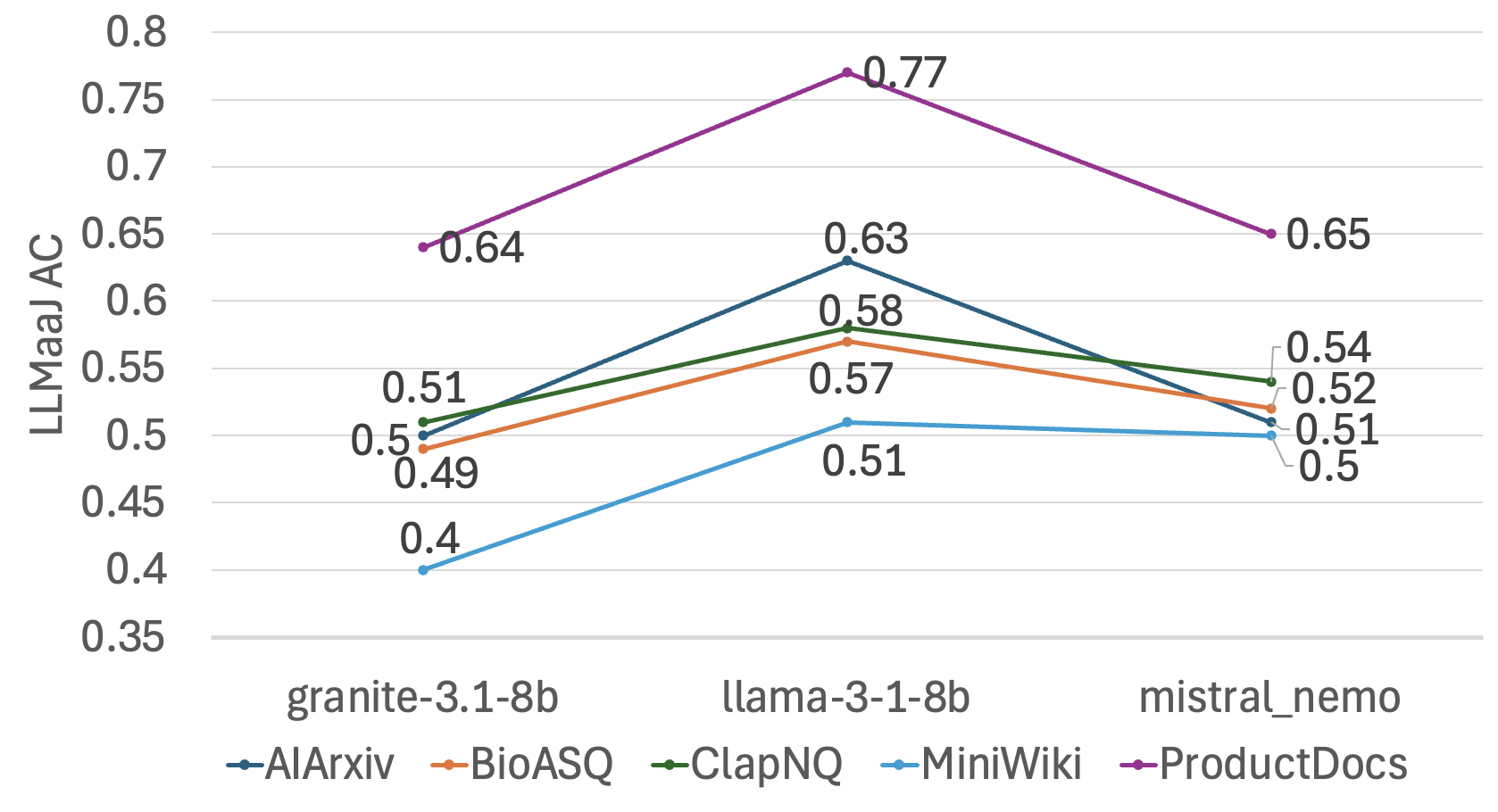}
    \label{fig:ragas_dataset_model_effect}}
  \hspace{0.02\textwidth}
  \subfigure[\lexical]{
    \centering
    
    \includegraphics[clip, trim=0.18cm 0.1cm 1.2cm 0cm, width=0.98\columnwidth]{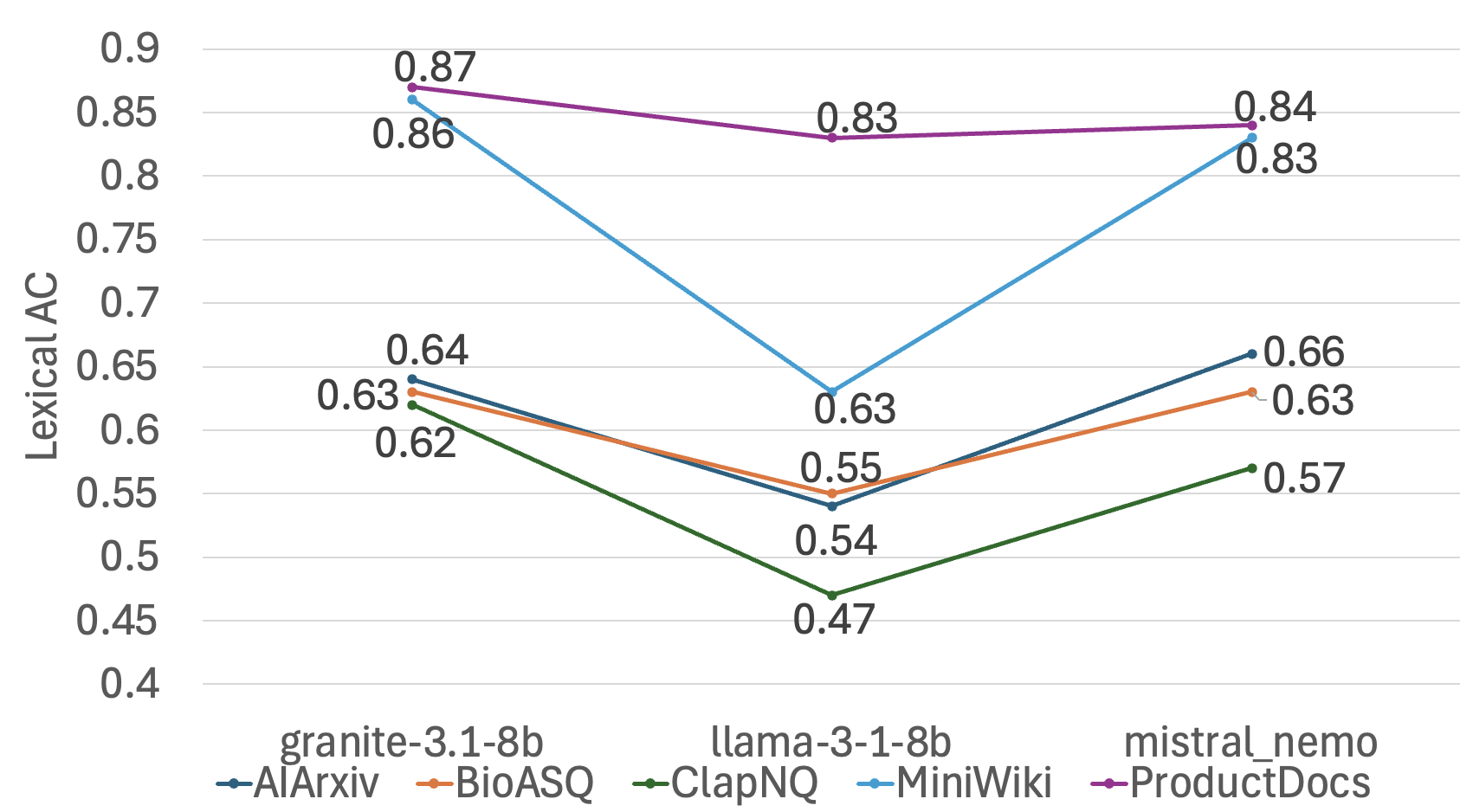}
    \label{fig:lexical_dataset_model_effect}
 }
  \caption{The effect of chosen optimization metric on the generative model within the best RAG configuration. Shown is the maximal answer correctness score per dataset and model (the highest of the $54$ configurations in which the model appears).}
  \label{fig:dataset_model_effect}
\end{figure*}

\begin{figure*}[!h]
  \centering
  %--- Subfigure (a): AC-RAGAS ---
  \subfigure[\llmaaj]{
    \centering
    % trim=left bottom right top
    \includegraphics[clip, trim=0.2cm 2.35cm 0.8cm 0cm, width=0.98\columnwidth, height=0.5\textheight]{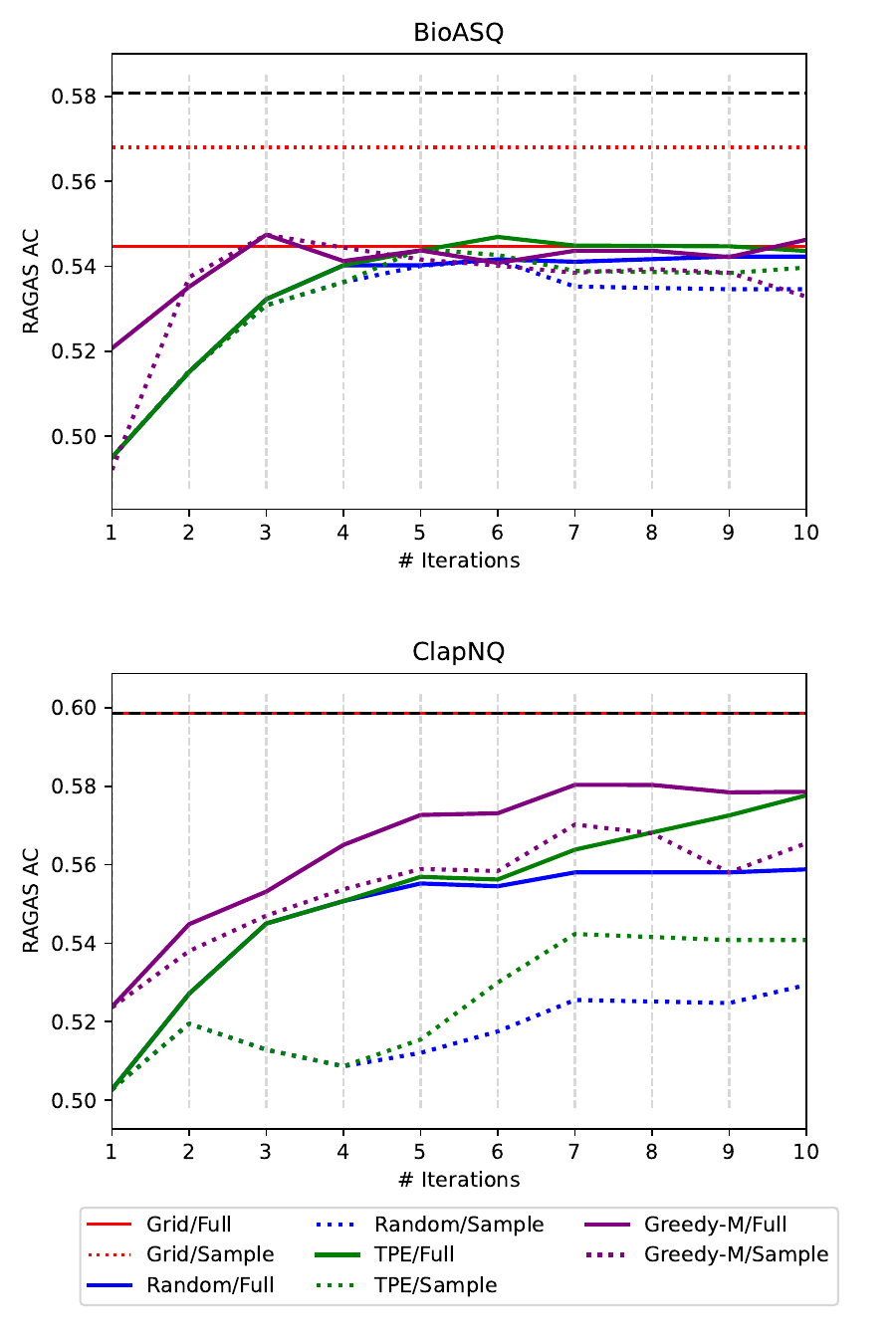}
    %\label{fig:ragas_sample_data_test_results}
  }
  \hspace{0.02\textwidth}
  %--- Subfigure (b): AC-lexical ---
  \subfigure[\lexical]{
    \centering
    \includegraphics[clip, trim=0.2cm 2.35cm 0.8cm 0cm, width=0.98\columnwidth, height=0.5\textheight]{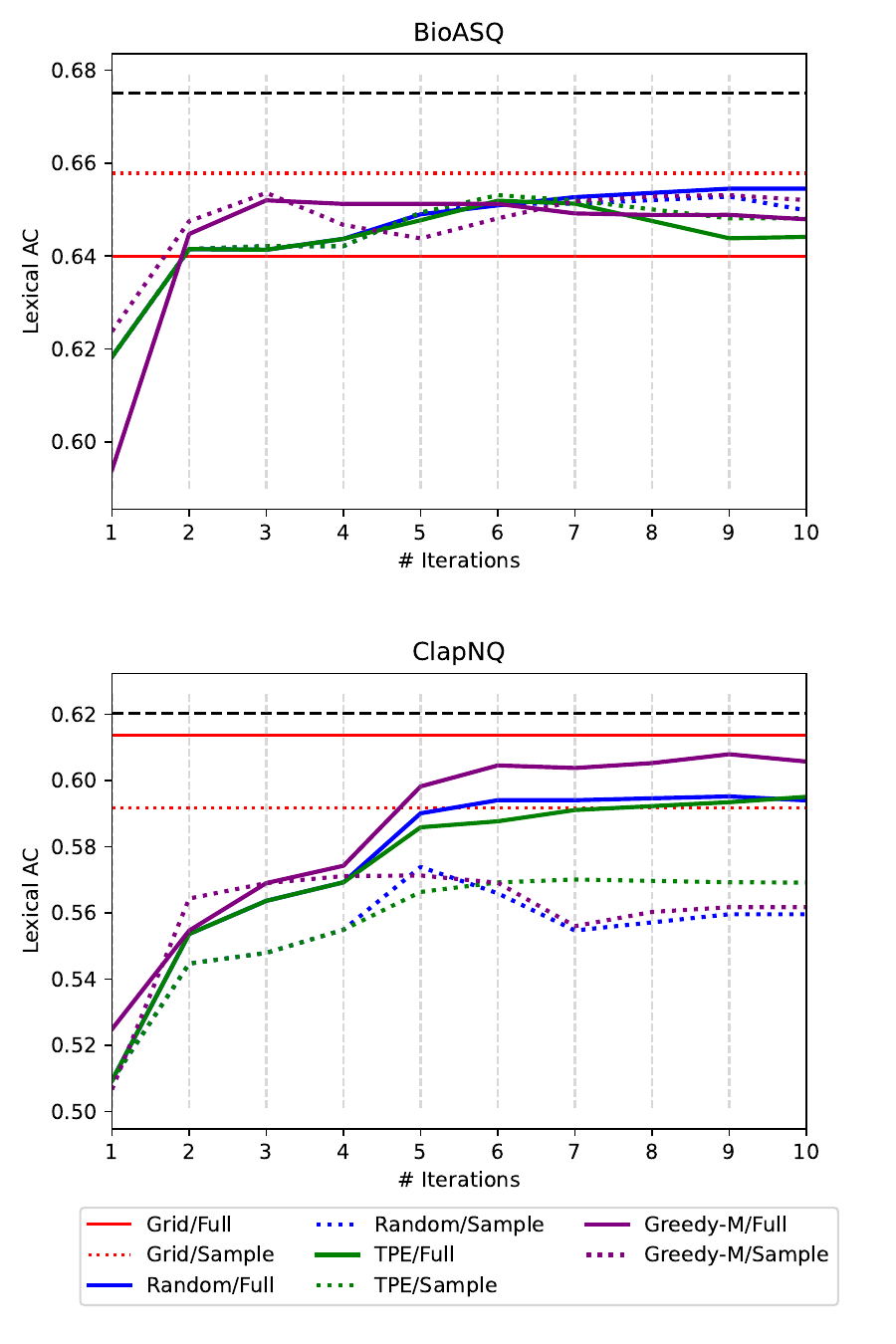}
    %\label{fig:lexical_sample_data_test_results}
  }
    \includegraphics[clip, trim=0cm 0cm 0cm 0cm, width=0.6\textwidth, height=0.07\textheight]{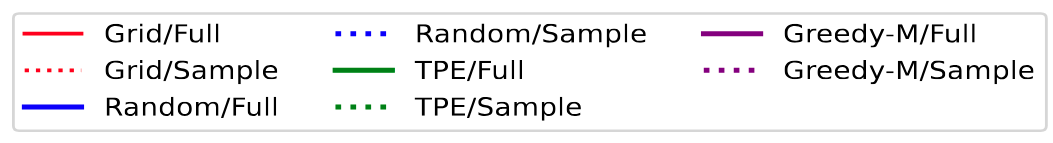}
  %--- Caption ---
  \caption{
    Per-iteration  performance on the test sets of the two largest datasets, for HPO algorithms optimized using the \textbf{full} development data (solid lines) or its \textbf{sample} (dotted).
    The dashed black lines show the best achievable test performance.
    The solid (dashed) red lines are the performance of the best configuration chosen by (sampled) development set evaluation.
  }
  \label{fig:sample_data_test_results}
\end{figure*}

The grid results serve to establish two important performance baselines.
The first is the performance of the best configuration selected directly from test set evaluation (dashed black lines in \figureRef{fig:full_data_test_results}).
The second is the best configuration chosen based on development set evaluation, and evaluated on the test set (red lines). 
The gap between these baselines reflects the inherent challenge of generalization. 
Since HPO algorithms operate solely on the development set, their realistic target is the second baseline.

The grid search results also reveal the impact of the different RAG pipeline parameters on the measured RAG performance. Overall, we see that almost all of the parameter choices have some effect on performance, however in our experiments the choice of generative model had the largest impact on the eventual answer correctness. For a detailed statistical analysis of the impact of the different RAG parameters, refer to Appendix~\ref{app:grid_search_stats}.

\subsection{HPO Results}
% \paragraph{HPO Results}

\figureRef{fig:full_data_test_results} presents the per-iteration performance of the HPO algorithms on the test sets, when optimizing for the lexical and LLMaaJ-based answer correctness metrics.\footnote{Answer faithfulness follow similar trends, see \supp \ref{app:additional_results}.}
Across all datasets and metrics, the results consistently show that exploring around $10$ configurations suffices to match the performance of a full grid search over all $162$ configurations. This is a strong result, demonstrating the robustness of HPO over diverse domains and evaluation metrics.

Also evident is that the difficulty of the optimization problem varies between datasets. 
For instance, in \clapnq convergence is rather slow. For \miniwiki, which is rich in good performing configurations (see \figureRef{fig:configurations_distribution_ragas}), finding a top configuration is easier, with an effective HPO algorithm such as \greedyModel, as few as three iterations can yield a top RAG configuration (by selecting the optimal generative model early). 
In contrast, even in this easy scenario, a method like \greedyRetrieverCC converges slower then the alternatives.
This underscores the importance of the HPO algorithm choice.

Among greedy methods, the order of parameter optimization is critical. The results show that algorithms starting with optimizing retrieval-related parameters first (\greedyRetriever and \greedyRetrieverCC) require more iterations to find good configurations. 

Interestingly, the naive option of random sampling also finds a good RAG configuration after a small number of iterations. 
That is likely due to the large impact of the generative model choice on performance, as reflected by the fast convergence achieved by the \greedyModel approach.

The complex \tpe algorithm was found to be roughly equivalent in quality to random choice. In larger or continuous spaces, \tpe may offer greater advantages.

% ADD FROM REBUTAL
% From our results and analyses, the most straightforward explanation for the success of simple methods is the large impact of the choice of generative model on the final results. This is reflected in the HPO results for the Greedy-M approach, which starts by choosing an LLM (see Figure 2). The impact of the model choice is also evident in the analysis of the performance of all configurations across the search space, as shown in Figure 5 – we see a clear separation between configurations based on the chosen model. Where most of the performance difference stems from this model choice, many simple methods can quickly converge to the most suitable LLM, and thus to a good “cluster” of high-performance configurations. We thank you for pointing this out, indeed this is an important point that we will clarify and discuss more explicitly in the text. 

%\section{Analysis}
%\label{sec:insights}
%Next, we analyze two aspects of our experiments: the impact of the chosen evaluation metric and cost.

\subsection{Impact of Metric Choice}
% \paragraph{Impact of Metric Choice} %\label{ssec:metric_impact}

The results of \greedym in \figureRef{fig:full_data_test_results} show performance is boosted significantly when the generative model parameter is optimized first, suggesting its importance. 
We therefore further investigate the best performing model in each setup.

\figureRef{fig:dataset_model_effect} reports the maximal answer correctness score per dataset for each generative model, computed as the highest of the $162/3=54$ configurations in which the model appears.
When optimizing by \llmaaj, the best RAG configuration consistently involves \lamma, whereas for \lexical, \granite or \mistral are better.
This difference stems from the nature of the metrics, as \lexical is recall-oriented while \llmaaj balances precision and recall. 
These findings emphasize the critical role of optimization objective selection. 
Optimal configurations can differ substantially depending on the chosen metric, and thus this choice should carefully align with the intended application.

\subsection{Cost Considerations}
% \paragraph{Cost Considerations} %
\label{ssec:costcons}
The cost of each HPO algorithm was tracked by counting the number of tokens embedded during indexing, and the number of tokens used in generation (input and output).
For each algorithm, we computed its total cost so far at a specific iteration, by accumulating these token counts over the   configurations evaluated up to that iteration (including).
The cost of indices used by multiple configurations was counted just once. 
Per-iteration plots of these counts are in \supp~\ref{app:tokens_results}.

Overall, generation costs were similar across algorithms and datasets.  
The embedding costs, dependent on the number of different indices created by the algorithm, behaved similarly across datasets with  variations between algorithms.
The \greedyr and \greedyrcc algorithms create a new index at each iteration until all retrieval parameters are optimized, making them initially expensive. 
Other algorithms like \tpe and \random lack a mechanism to favor index reuse. 
\greedym begins by optimizing the generative model using a single index, making it cost-efficient when budget constraints or iteration limits are tight.

\subsection{Efficient HPO}
The results of \figureRef{fig:full_data_test_results} were obtained with each RAG configuration evaluated on the whole development set.
While simple, this option is costly for large datasets. 
Prior work in other domains suggests that random sampling of evaluation benchmarks can reduce costs without sacrificing evaluation quality \cite{perlitz-etal-2024-efficient, tinyBenchmarks2024}, and we therefore explore this direction.
%hus, we experiment with HPO that uses a sample of the development set.
To our knowledge, this is the first study of that direction in the context of HPO for RAG. 

To adapt sampling to RAG, we sample both the benchmark and the underlying corpus.
% to reduce both inference and embedding costs. 
Specifically, focusing on the larger datasets of \bioasq and \clapnq, 10\% of the development QA pairs were sampled along with their corresponding gold documents (those containing the answers to the sampled questions).
To preserve realistic retrieval conditions we add ``noise'' -- documents not containing an answer to any of the sampled questions -- at a ratio of $9$ such documents per gold document. 
This yields $100$ sampled benchmark questions per dataset, with the sampled corpora comprising of $1$K (i.e. $1000$) documents for \clapnq (out of $178$K), and $10$K for \bioasq (out of $40$K).\footnote{\bioasq has multiple gold documents per question, which yields more sampled documents.}

Following sampling, we repeat the  experiments using the best HPO methods: \random, \tpe, and \greedym. 
\figureRef{fig:sample_data_test_results} compares performance when optimizing for \llmaaj and \lexical, using the full (solid lines) or sampled (dotted) development sets. 
For \bioasq sampling has a negligible impact. 
For \clapnq, results differ per algorithm and metric, with a suboptimal configuration identified by \tpe and \random. For \greedym and the \llmaaj metric, the performance drop is moderate. 

With this approach, cost reductions are substantial.
Inference costs for a given configuration are $10$x cheaper, as $10$\% of the questions are used. 
Indexing costs drop by $4$x for \bioasq and $178$x for \clapnq. 
These savings make sampling highly attractive for HPO over large datasets.

In summary, development set sampling offers a promising path towards efficient RAG HPO. Combined with the \greedym approach, the trade-off between cost and performance remains favorable, making it a practical choice for real-world applications.

%Improved sampling approaches might be more robust to the choice of the HPO algorithm. We leave this for future work.

%\input{figures/dev_retrieval_fig}

%\paragraph{Impact of Retrieval Quality} %\label{ssec:retrieval}
%\figureRef{fig:retrieval} depicts the relationship between context correctness and answer correctness.
%As expected, the two measures are correlated -- when retrieval quality is low, so is answer quality.
%Also evident is the  influence of the generative model choice on answer quality.

%The impact of context correctness on answer correctness diminishes as context correctness increases, as implied by the line slopes in \figureRef{fig:retrieval} being steeper at low context correctness values. This behavior can be attributed to the fact that several retrieved chunks are utilized during generation.
%When context correctness is low, there is a high probability that none of the chunks are relevant, and the generated answer will therefore be incorrect.
%Conversely, when at least one relevant chunk is retrieved,  the quality of the answer will largely rely on generation hyper-parameters.  

\section{Conclusion}

We presented a comprehensive study of HPO  for RAG in a generalization setup that reflects real-world usage.
Our evaluation spans five HPO algorithms, three evaluations metrics, and multiple datasets from diverse domains.
One is a newly curated enterprise product documentation dataset, released as part of this work, for use by the community.

Our findings are that HPO systematically boosts RAG performance significantly.
Compared to an arbitrarily chosen RAG configuration, running $10$ HPO iterations yield gains of up to $20$\% (see \figureRef{fig:full_data_test_results}, comparing the performance at the first and last iterations).  
While our experiments focus on core RAG components, the potential impact on more complex systems parametrized by larger search spaces is likely even greater. 
For example, exploring HPO in the context of multi-modal or agentic RAG pipelines seems a promising direction for future work. 

%% Efficiency 

We showed that RAG HPO can be performed efficiently.
Even without prior knowledge of the RAG pipeline parameters, exploring a small subset of the configuration space is often sufficient.
Simple strategies such as random sampling perform surprisingly well, while a greedy approach that prioritizes model selection outperforms the common practice of sequential optimization by pipeline order.

Our results highlight the importance of the optimization objective choice, as different objective choices lead to different optimal RAG configurations. 
We further show that development set sampling can reduce the costs of HPO for RAG by orders of magnitude.
With the use of the \greedyModel algorithm, the saved compute, at the mild cost of performance, may be attractive for many users.

For practitioners interested in boosting the performance of their RAG pipelines, we strongly suggest the use of HPO, and offer the following recommendations. 
Carefully choose an optimization metric that reflects the goals of the application.
With that, evaluate multiple configurations with randomly picked parameter values, this initial quick exploration is likely to give valuable gains.  
Next, improve efficiency by using  a greedy HPO algorithm that optimizes model choices first, that will provide faster convergence.
For large datasets, combine that algorithm with development set sampling to efficiently find a top-performing RAG configuration.

Finally, we open-source our complete grid search results over the development and test sets for all datasets.\footnote{\url{https://github.com/IBM/rag-hpo-bench}}
To our knowledge, we are the first to release such a resource for RAG.
Building on these results, further research can explore new HPO techniques without incurring the substantial cost of running many RAG configuration across datasets. 
Our release also includes the generation outputs for each configuration, enabling easy computation of additional metrics and their analysis in the context of HPO. 
We hope this release will serve as a valuable contribution to research on HPO for RAG.

%\section*{Limitations}
%We limited the size of our search space to enable a full grid search on all  datasets in a reasonable time and cost. 
%Further work could explore a larger search space without comparing to a full grid search. 

%Our experiments optimized a simple linear RAG workflow comprising retrieval followed by generation. Additional RAG approaches may introduce additional hyper-parameters to the mix, such as  filters, rerankers, answer reviewers, HyDE etc.
%Moreover, our prompts were fixed per generation model and we did not explore prompt optimization. 

%Our analysis considers only English textual datasets with answerable questions. The comparison of performance for multi-modal, multi-lingual datasets that include also unanswerable questions is left for future work.

%It should be noted that our method for retrieval quality evaluation is not accurate: any chunk from the gold document is considered a correct prediction even if it does not include the answer to the question. 

%\bibliography{custom}

% We emphasize the need for an automated approach. 
%For example, the impact of the optimization metric on the optimal RAG configuration (see \figureRef{fig:dataset_model_effect}) dictates that when changing metrics even the same RAG pipeline must be re-adjusted to the new metric with HPO. Similarly, changes in available models or evolving datasets require revising the chosen RAG configuration.

% \section{Acknowledgments}

\bibliography{main}

@article{lindauer2022smac3,
  author  = {Marius Lindauer and Katharina Eggensperger and Matthias Feurer and André Biedenkapp and Difan Deng and Carolin Benjamins and Tim Ruhkopf and René Sass and Frank Hutter},
  title   = {SMAC3: A Versatile Bayesian Optimization Package for Hyperparameter Optimization},
  journal = {Journal of Machine Learning Research},
  year    = {2022},
  volume  = {23},
  number  = {54},
  pages   = {1--9},
  url     = {http://jmlr.org/papers/v23/21-0888.html}
}

@inproceedings{falkner2018bohb,
  title={BOHB: Robust and efficient hyperparameter optimization at scale},
  author={Falkner, Stefan and Klein, Aaron and Hutter, Frank},
  booktitle={International conference on machine learning},
  pages={1437--1446},
  year={2018},
  organization={PMLR}
}

@article{barker2025multi_objective_optimization,
  title={Faster, Cheaper, Better: Multi-Objective Hyperparameter Optimization for LLM and RAG Systems},
  author={Barker, Matthew and Bell, Andrew and Thomas, Evan and Carr, James and Andrews, Thomas and Bhatt, Umang},
  journal={arXiv preprint arXiv:2502.18635},
  year={2025}
}

@inproceedings{smith2008question,
  title={Question generation as a competitive undergraduate course project},
  author={Smith, Noah A and Heilman, Michael and Hwa, Rebecca},
  booktitle={Proceedings of the NSF Workshop on the Question Generation Shared Task and Evaluation Challenge},
  volume={9},
  year={2008}
}

@inproceedings{perlitz-etal-2024-efficient,
    title = "Efficient Benchmarking (of Language Models)",
    author = "Perlitz, Yotam  and
      Bandel, Elron  and
      Gera, Ariel  and
      Arviv, Ofir  and
      Ein-Dor, Liat  and
      Shnarch, Eyal  and
      Slonim, Noam  and
      Shmueli-Scheuer, Michal  and
      Choshen, Leshem",
    editor = "Duh, Kevin  and
      Gomez, Helena  and
      Bethard, Steven",
    booktitle = "Proceedings of the 2024 Conference of the North American Chapter of the Association for Computational Linguistics: Human Language Technologies (Volume 1: Long Papers)",
    month = jun,
    year = "2024",
    address = "Mexico City, Mexico",
    publisher = "Association for Computational Linguistics",
    url = "https://aclanthology.org/2024.naacl-long.139/",
    doi = "10.18653/v1/2024.naacl-long.139",
    pages = "2519--2536"
}

@inproceedings{tinyBenchmarks2024,
author = {Polo, Felipe Maia and Weber, Lucas and Choshen, Leshem and Sun, Yuekai and Xu, Gongjun and Yurochkin, Mikhail},
title = {tinyBenchmarks: evaluating LLMs with fewer examples},
year = {2024},
publisher = {JMLR.org},
booktitle = {Proceedings of the 41st International Conference on Machine Learning},
articleno = {1396},
numpages = {24},
location = {Vienna, Austria},
series = {ICML'24}
}

@article{eibich2024aragog,
  title={ARAGOG: Advanced RAG Output Grading},
  author={Eibich, Matou{\v{s}} and Nagpal, Shivay and Fred-Ojala, Alexander},
  journal={arXiv preprint arXiv:2404.01037},
  year={2024}
}

@article{kwiatkowski2019natural,
  title={Natural questions: a benchmark for question answering research},
  author={Kwiatkowski, Tom and Palomaki, Jennimaria and Redfield, Olivia and Collins, Michael and Parikh, Ankur and Alberti, Chris and Epstein, Danielle and Polosukhin, Illia and Devlin, Jacob and Lee, Kenton and others},
  journal={Transactions of the Association for Computational Linguistics},
  volume={7},
  pages={453--466},
  year={2019},
  publisher={MIT Press One Rogers Street, Cambridge, MA 02142-1209, USA journals-info~…}
}

@article{kim2024autorag,
  title={AutoRAG: Automated Framework for optimization of Retrieval Augmented Generation Pipeline},
  author={Kim, Dongkyu and Kim, Byoungwook and Han, Donggeon and Eibich, Matou{\v{s}}},
  journal={arXiv preprint arXiv:2410.20878},
  year={2024}
}

@misc{yehudai2024,
      title={Genie: Achieving Human Parity in Content-Grounded Datasets Generation}, 
      author={Asaf Yehudai and Boaz Carmeli and Yosi Mass and Ofir Arviv and Nathaniel Mills and Assaf Toledo and Eyal Shnarch and Leshem Choshen},
      year={2024},
      eprint={2401.14367},
      archivePrefix={arXiv},
      primaryClass={cs.CL},
      url={https://arxiv.org/abs/2401.14367}, 
}

@inproceedings{fu-etal-2024-autorag,
    title = "{A}uto{RAG}-{HP}: Automatic Online Hyper-Parameter Tuning for Retrieval-Augmented Generation",
    author = "Fu, Jia  and
      Qin, Xiaoting  and
      Yang, Fangkai  and
      Wang, Lu  and
      Zhang, Jue  and
      Lin, Qingwei  and
      Chen, Yubo  and
      Zhang, Dongmei  and
      Rajmohan, Saravan  and
      Zhang, Qi",
    editor = "Al-Onaizan, Yaser  and
      Bansal, Mohit  and
      Chen, Yun-Nung",
    booktitle = "Findings of the Association for Computational Linguistics: EMNLP 2024",
    month = nov,
    year = "2024",
    address = "Miami, Florida, USA",
    publisher = "Association for Computational Linguistics",
    url = "https://aclanthology.org/2024.findings-emnlp.223",
    doi = "10.18653/v1/2024.findings-emnlp.223",
    pages = "3875--3891",
}

@misc{ragas,
      title={RAGAS: Automated Evaluation of Retrieval Augmented Generation}, 
      author={Shahul Es and Jithin James and Luis Espinosa-Anke and Steven Schockaert},
      year={2023},
      eprint={2309.15217},
      archivePrefix={arXiv},
      primaryClass={cs.CL},
      url={https://arxiv.org/abs/2309.15217}, 
}

@article{krithara2023bioasq,
  title={BioASQ-QA: A manually curated corpus for Biomedical Question Answering},
  author={Krithara, Anastasia and Nentidis, Anastasios and Bougiatiotis, Konstantinos and Paliouras, Georgios},
  journal={Scientific Data},
  volume={10},
  number={1},
  pages={170},
  year={2023},
  publisher={Nature Publishing Group UK London}
}

@inproceedings{voorhees-tice-2000-trec,
    title = "The {TREC}-8 Question Answering Track",
    author = "Voorhees, Ellen M.  and
      Tice, Dawn M.",
    editor = "Gavrilidou, M.  and
      Carayannis, G.  and
      Markantonatou, S.  and
      Piperidis, S.  and
      Stainhauer, G.",
    booktitle = "Proceedings of the Second International Conference on Language Resources and Evaluation ({LREC}`00)",
    month = may,
    year = "2000",
    address = "Athens, Greece",
    publisher = "European Language Resources Association (ELRA)",
    url = "https://aclanthology.org/L00-1018/"
}

@article{adlakha-etal-2024-evaluating,
    title = "Evaluating Correctness and Faithfulness of Instruction-Following Models for Question Answering",
    author = "Adlakha, Vaibhav  and
      BehnamGhader, Parishad  and
      Lu, Xing Han  and
      Meade, Nicholas  and
      Reddy, Siva",
    journal = "Transactions of the Association for Computational Linguistics",
    volume = "12",
    year = "2024",
    address = "Cambridge, MA",
    publisher = "MIT Press",
    url = "https://aclanthology.org/2024.tacl-1.38/",
    doi = "10.1162/tacl_a_00667",
    pages = "681--699"
}

@misc{rosenthal2024clapnq,
      title={CLAPNQ: Cohesive Long-form Answers from Passages in Natural Questions for RAG systems}, 
      author={Sara Rosenthal and Avirup Sil and Radu Florian and Salim Roukos},
      year={2024},
      eprint={2404.02103},
      archivePrefix={arXiv},
      primaryClass={cs.CL}
}

@misc{multilingual-e5-large,
    title = {Multilingual E5 Text Embeddings: A Technical Report},
    author = {Liang Wang and Nan Yang and Xiaolong Huang and Linjun Yang and Rangan Majumderand Furu Wei},
    year = {2024},
    eprint = {2402.05672},
    archivePrefix={arXiv},
    primaryClass={cs.CL}
}

@misc{bge_embedding,
      title={C-Pack: Packaged Resources To Advance General Chinese Embedding}, 
      author={Shitao Xiao and Zheng Liu and Peitian Zhang and Niklas Muennighoff},
      year={2023},
      eprint={2309.07597},
      archivePrefix={arXiv},
      primaryClass={cs.CL}
}

@misc{meta2024llama3.1,
  title = {Introducing Llama 3.1: Our most capable models to date},
  author = {Meta AI},
  year = {2024},
  url = {https://ai.meta.com/blog/meta-llama-3-1/}
}

@misc{ibm2024graniteembedding125m,
  title = {Granite-Embedding-125M-English Model Card},
  author = {IBM},
  year = {2024},
  url = {https://huggingface.co/ibm-granite/granite-embedding-125m-english}
}

@misc{mistral2024nemo,
  title = {Mistral NeMo: A state-of-the-art 12B model with 128k context length},
  author = {Mistral AI and NVIDIA},
  year = {2024},
  url = {https://mistral.ai/news/mistral-nemo/}
}

@misc{openai2024gpt4omini,
  title = {GPT-4o mini: Advancing Cost-Efficient Intelligence},
  author = {OpenAI},
  year = {2024},
  url = {https://openai.com/index/gpt-4o-mini-advancing-cost-efficient-intelligence/}
}

@misc{granite2024,
  author       = {Granite Team, IBM},
  title        = {Granite 3.0 Language Models},
  year         = {2024},
  url          = {https://github.com/ibm-granite/granite-3.0-language-models/blob/main/paper.pdf},
  note         = {Accessed: 2025-02-14}
}

@article{liaw2018tune,
    title={Tune: A Research Platform for Distributed Model Selection and Training},
    author={Liaw, Richard and Liang, Eric and Nishihara, Robert
            and Moritz, Philipp and Gonzalez, Joseph E and Stoica, Ion},
    journal={arXiv preprint arXiv:1807.05118},
    year={2018}
}

@article{optuna,
  author    = {Takuya Akiba and Satoshi Sano and Tatsuya Koyama and Yuji Matsumoto and Masanori Ohta},
  title     = {Optuna: A Next-generation Hyperparameter Optimization Framework},
  journal   = {arXiv preprint arXiv:1907.10902},
  year      = {2019},
  url       = {https://arxiv.org/abs/1907.10902}
}

@inproceedings{hyperopt,
  author    = {Bergstra, J. and Yamins, D. and Cox, D. D.},
  title     = {Making a Science of Model Search: Hyperparameter Optimization in Hundreds of Dimensions for Vision Architectures},
  booktitle = {Proceedings of the 30th International Conference on Machine Learning (ICML 2013)},
  pages     = {I-115--I-123},
  year      = {2013},
  url       = {http://proceedings.mlr.press/v28/bergstra13.pdf}
}

@article{crud-rag,
  author = {Yuanjie Lyu and Zhiyu Li and Simin Niu and Feiyu Xiong and Bo Tang and Wenjin Wang and Hao Wu and Huanyong Liu and Tong Xu and Enhong Chen},
  title = {CRUD-RAG: A Comprehensive Chinese Benchmark for Retrieval-Augmented Generation of Large Language Models},
  journal = {arXiv:2401.17043 [cs.CL]},
  year = {2024},
  url = {https://doi.org/10.48550/arXiv.2401.17043},
  note = {Accessed: 2025-02-14},
}

@misc{tpe,
      title={Tree-Structured Parzen Estimator: Understanding Its Algorithm Components and Their Roles for Better Empirical Performance}, 
      author={Shuhei Watanabe},
      year={2023},
      eprint={2304.11127},
      archivePrefix={arXiv},
      primaryClass={cs.LG},
      url={https://arxiv.org/abs/2304.11127}, 
}

@inproceedings{milvus,
author = {Wang, Jianguo and Yi, Xiaomeng and Guo, Rentong and Jin, Hai and Xu, Peng and Li, Shengjun and Wang, Xiangyu and Guo, Xiangzhou and Li, Chengming and Xu, Xiaohai and Yu, Kun and Yuan, Yuxing and Zou, Yinghao and Long, Jiquan and Cai, Yudong and Li, Zhenxiang and Zhang, Zhifeng and Mo, Yihua and Gu, Jun and Jiang, Ruiyi and Wei, Yi and Xie, Charles},
title = {Milvus: A Purpose-Built Vector Data Management System},
year = {2021},
isbn = {9781450383431},
publisher = {Association for Computing Machinery},
address = {New York, NY, USA},
url = {https://doi.org/10.1145/3448016.3457550},
doi = {10.1145/3448016.3457550},
booktitle = {Proceedings of the 2021 International Conference on Management of Data},
pages = {2614–2627},
numpages = {14},
location = {Virtual Event, China},
series = {SIGMOD '21}
}

@misc{hnsw,
      title={Efficient and robust approximate nearest neighbor search using Hierarchical Navigable Small World graphs}, 
      author={Yu. A. Malkov and D. A. Yashunin},
      year={2018},
      eprint={1603.09320},
      archivePrefix={arXiv},
      primaryClass={cs.DS},
      url={https://arxiv.org/abs/1603.09320}, 
}

@misc{ragbench,
      title={RAGEval: Scenario Specific RAG Evaluation Dataset Generation Framework}, 
      author={Kunlun Zhu and Yifan Luo and Dingling Xu and Ruobing Wang and Shi Yu and Shuo Wang and Yukun Yan and Zhenghao Liu and Xu Han and Zhiyuan Liu and Maosong Sun},
      year={2024},
      eprint={2408.01262},
      archivePrefix={arXiv},
      primaryClass={cs.CL},
      url={https://arxiv.org/abs/2408.01262}, 
}

@inproceedings{wang2024searching,
  title={Searching for best practices in retrieval-augmented generation},
  author={Wang, Xiaohua and Wang, Zhenghua and Gao, Xuan and Zhang, Feiran and Wu, Yixin and Xu, Zhibo and Shi, Tianyuan and Wang, Zhengyuan and Li, Shizheng and Qian, Qi and others},
  booktitle={Proceedings of the 2024 Conference on Empirical Methods in Natural Language Processing},
  pages={17716--17736},
  year={2024}
}

@inproceedings{rag1,
author = {Lewis, Patrick and Perez, Ethan and Piktus, Aleksandra and Petroni, Fabio and Karpukhin, Vladimir and Goyal, Naman and K\"{u}ttler, Heinrich and Lewis, Mike and Yih, Wen-tau and Rockt\"{a}schel, Tim and Riedel, Sebastian and Kiela, Douwe},
title = {Retrieval-augmented generation for knowledge-intensive NLP tasks},
year = {2020},
isbn = {9781713829546},
publisher = {Curran Associates Inc.},
address = {Red Hook, NY, USA},
booktitle = {Proceedings of the 34th International Conference on Neural Information Processing Systems},
articleno = {793},
numpages = {16},
location = {Vancouver, BC, Canada},
series = {NIPS '20}
}

@misc{rag4,
      title={Searching for Best Practices in Retrieval-Augmented Generation}, 
      author={Xiaohua Wang and Zhenghua Wang and Xuan Gao and Feiran Zhang and Yixin Wu and Zhibo Xu and Tianyuan Shi and Zhengyuan Wang and Shizheng Li and Qi Qian and Ruicheng Yin and Changze Lv and Xiaoqing Zheng and Xuanjing Huang},
      year={2024},
      eprint={2407.01219},
      archivePrefix={arXiv},
      primaryClass={cs.CL},
      url={https://arxiv.org/abs/2407.01219}, 
}

@misc{rag3,
      title={A Survey on Retrieval-Augmented Text Generation for Large Language Models}, 
      author={Yizheng Huang and Jimmy Huang},
      year={2024},
      eprint={2404.10981},
      archivePrefix={arXiv},
      primaryClass={cs.IR},
      url={https://arxiv.org/abs/2404.10981}, 
}

@misc{rag2,
      title={Retrieval-Augmented Generation for Large Language Models: A Survey}, 
      author={Yunfan Gao and Yun Xiong and Xinyu Gao and Kangxiang Jia and Jinliu Pan and Yuxi Bi and Yi Dai and Jiawei Sun and Meng Wang and Haofen Wang},
      year={2024},
      eprint={2312.10997},
      archivePrefix={arXiv},
      primaryClass={cs.CL},
      url={https://arxiv.org/abs/2312.10997}, 
}

\appendix

\section{\watsonxdataset additional details}
\label{sec:appendix-datasets}
\label{app:watsonx_details}

%Example questions per dataset are listed in \tableRef{app:dataset_examples}.
%\subsection{}

% \input{tables/appendix-datasets}

As stated in the body of the paper, the \watsonxdataset benchmark includes $75$ QA pairs and gold document labels, of which $50$ were generated synthetically. The benchmark contains five fields: a question, a gold answer and, for computing context relevance metrics, the gold passage id and its content.  Some example benchmark entries are given in \figureRef{fig:product_sample}. Synthetic generation was operated using falcon-180b model, and then manually filtered and reviewed for quality - the methodology we used is detailed in~\citet{yehudai2024}.

The prompt used for the synthetic generation is detailed in ~\figureRef{fig:synthetic_prompt}.

\section{Search Space Selection}
\label{sec:search_space_selection}
The search space values for the parameters  \chunksize, \chunkoverlap and \topk were chosen based on previous works:

\paragraph{Chunk size} \citet{wang2024searching} used the values $\{128, 256, 512, 1024, 2048\}$. They reported that the values $256$ and $512$ performed well (see Table~3 in \citet{wang2024searching}) in terms of faithfulness and relevancy. \citet{crud-rag} used $\{64, 128, 256, 512\}$. We chose three values $\{256, 384, 512\}$.
    
\paragraph{\topk} \citet{fu-etal-2024-autorag} experimented with $\{1, 3, 5, 7, 9\}$, and \citet{crud-rag} with $\{2, 4, 6, 8, 10\}$. We chose to use $\{3, 5, 10\}$.

\paragraph{\chunkoverlap} \citet{crud-rag} used $\{0\%, 10\%, 30\%, 50\%, 70\%\}$. Others have not considered this parameter. We used $\{0\%, 25\%\}$.

%fu-etal-2024-autorag
%R1: Findings of the Association for Computational Linguistics: EMNLP 2024, pages 3875–3891
%November 12-16, 2024 ©2024 Association for Computational LinguisticsAutoRAG-HP: Automatic Online Hyper-Parameter Tuning for
%Retrieval-Augmented Generation

%R2: Searching for Best Practices in Retrieval-Augmented
%Generation

%R3: CRUD-RAG: A Comprehensive Chinese Benchmark for
%Retrieval-Augmented Generation of Large Language Models

\section{Impact of specific parameters}
\label{app:grid_search_stats}

To test the impact of the different RAG pipeline parameter choices, we conduct statistical analyses over the grid-search results for each dataset.

Specifically, we fit a linear mixed-effects model on the dataset results, where the per-example answer correctness is the dependent variable, the choice of generative model, embedding model, chunk size, chunk overlap and retrieved K are the fixed effects we test for, and the individual examples are modeled as a random effect. In addition to the main effects, we include in our model possible \textit{interaction} effects: between the embedding model and the generative model, between the embedding model and the chunk size, between the chunk size and chunk overlap, and between the generative model and the retrieved K.

To assess the significance of each of the tested main effects and interactions, we performed likelihood ratio tests, comparing the full mixed-effects model to a reduced model that excludes a specific main effect or interaction. We report the resulting test statistics $\chi^2$ and significance values $p$ for each dataset in Tables \ref{tab:lrt_a}-\ref{tab:lrt_j}. As can be seen in the tables, most of the effects and interactions are statistically significant ($p < .05$), indicating that these choices do indeed affect the pipeline result. Consistently, the choice of the generative model has a particularly large effect, explaining much of the variance of the statistical model.

In addition, for each dataset and metric we report the marginal means for each chosen pipeline parameter, and their delta from the overall mean metric result. As can be seen in Tables \ref{tab:mmeans_a}-\ref{tab:mmeans_e}, the largest differences in the metric scores relate to the choice of generative model. In addition, the relative success of the different generative models depends on the chosen metric, as also shown in \figureRef{fig:dataset_model_effect}.

We conduct all analyses using the \emph{statsmodels} python library (v0.14.6). Parameters and marginal means were estimated using Restricted Maximum Likelihood (REML). For the likelihood-ratio-testing of effect significance, models were compared using Maximum Likelihood (ML).

\section{Additional results}
\label{app:additional_results}

\begin{figure}[t]
  \centering
  \includegraphics[width=1\columnwidth]{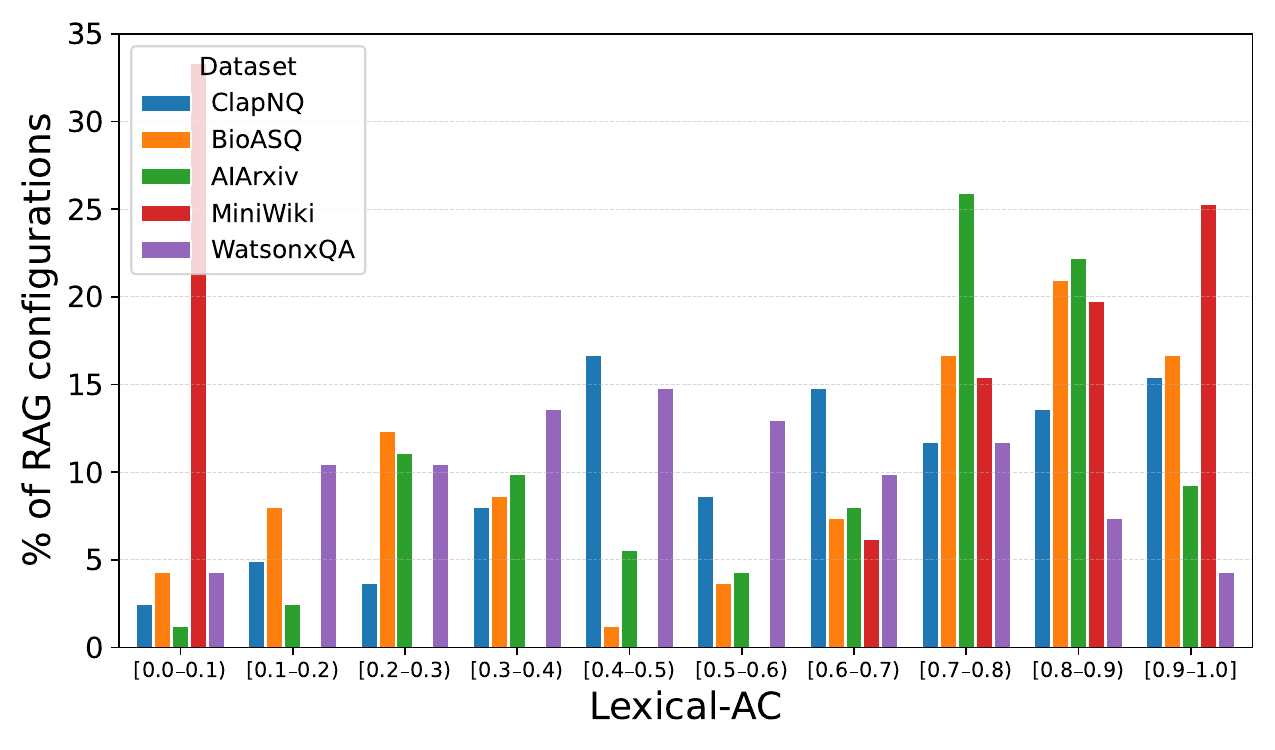}
  \vspace{1em}
  \includegraphics[width=1\columnwidth]{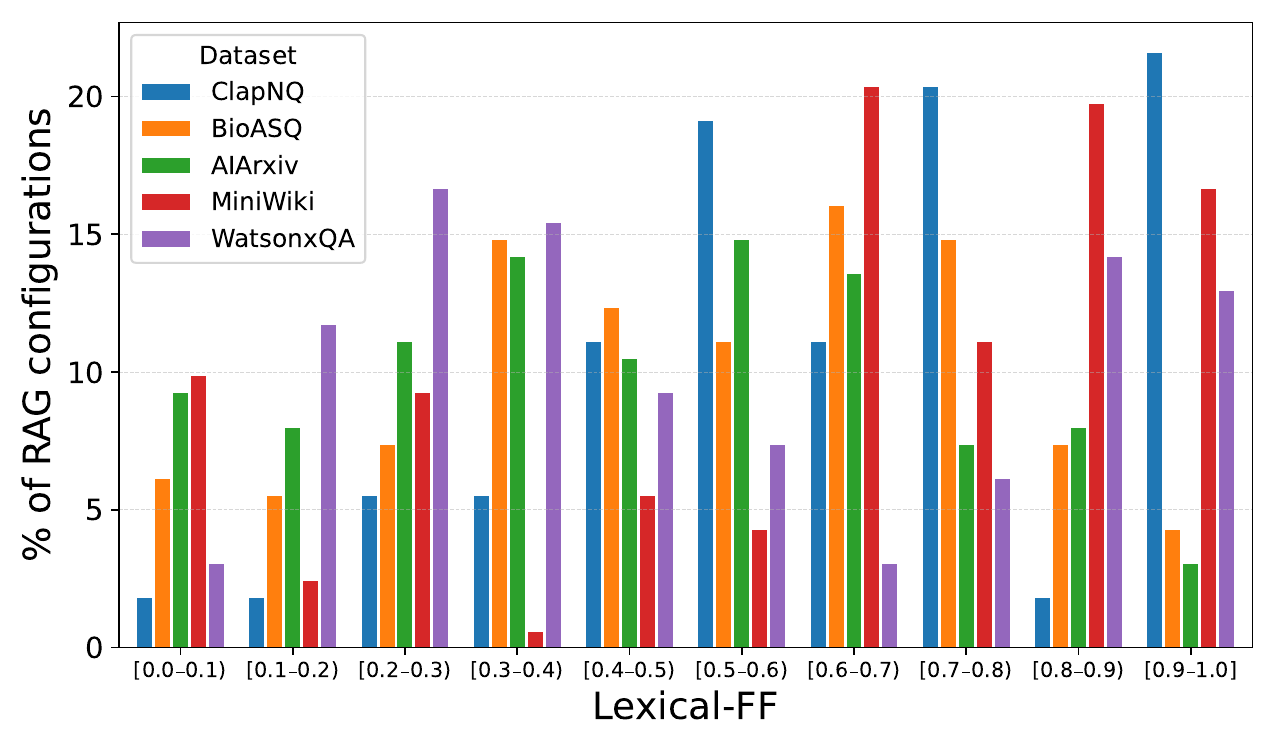}
  \caption{The percentage of RAG configurations assigned to each bin of normalized metric scores (with the \lexical or \faithfulness metrics), on the dev sets.}
  \label{fig:configurations_distribution_other}
\end{figure}

\begin{figure}[h!]
\centering
% left bottom right top
\includegraphics[clip, trim=0cm 0cm 0cm 0cm, width=0.95\columnwidth]{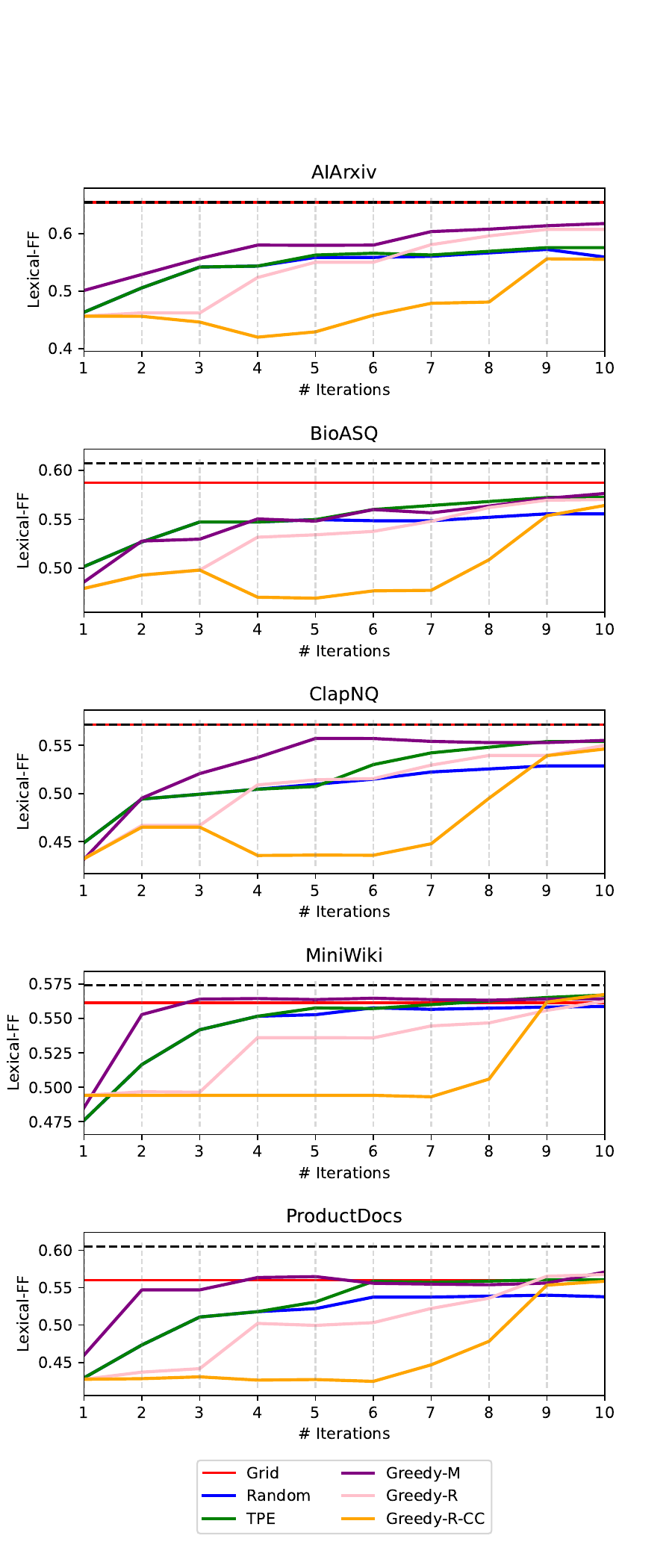}\\
\caption{Per-iteration performance of all HPO algorithms on the test sets of five datasets, optimizing answer faithfulness. The dashed black lines denote the best achievable performance on each test set. See \sectionRef{sec:results} for a discussion of the main results.}
\label{fig:full_data_test_results_faithfulness}
\end{figure}

See \sectionRef{sec:results} for a discussion of the main results.
\begin{itemize}
  \item Table~\ref{app:confidenceinterval} shows  
      the worst and best per-dataset \faithfulness metric scores on the development set. 
  \item \figureRef{fig:configurations_distribution_other} depicts the percentage of good and bad configurations for the five datasets and the \lexical and \faithfulness metrics.
  \item \figureRef{fig:full_data_test_results_faithfulness} details HPO results for the \faithfulness metric.

      %demonstrates statistical significance for the \faithfulness metric.
  
%  \item \figureRef{fig:lexical_sample_data_test_results} details development sampling results for the \lexical metric. 
\end{itemize}

\begin{table}
  \centering
  \begin{tabular}
  {
  >{\raggedright\arraybackslash}p{0.2\columnwidth}
  >{\centering\arraybackslash}p{0.13\columnwidth}
  >{\centering\arraybackslash}p{0.13\columnwidth}
  >{\centering\arraybackslash}p{0.13\columnwidth}
  }
    \toprule
    & \multicolumn{3}{c}{\textbf{\faithfulness}} \\
    \cmidrule(lr){2-4}
    \textbf{Dataset} & \textbf{Worst} & \textbf{Best} & \textbf{SE}\\
    \midrule
    \aiarxiv        & 0.28 & 0.64 & 0.03 \\
    \bioasq         & 0.38 & 0.60 & 0.01 \\
    \miniwiki       & 0.39 & 0.56 & 0.01 \\
    \clapnq         & 0.28 & 0.56 & 0.01 \\
    \watsonxdataset & 0.37 & 0.65 & 0.03 \\
    \bottomrule
  \end{tabular}
  \caption{\label{app:confidenceinterval}
    \textbf{Worst} and \textbf{Best} configuration scores per dataset on the development set for the \faithfulness metric. Also shown is the maximum standard error (\textbf{SE}) observed across all configurations.
  }
\end{table}

\section{Embedding and Generation Costs }
\label{app:tokens_results}

\begin{figure*}
    \centering
    % First Subfigure
    \subfigure[Total number of tokens from chunks sent to the embedding models.]{
        \centering
        \begin{tabular}{ccc}
            \includegraphics[trim=1cm 0 0 0, width=0.33\textwidth]{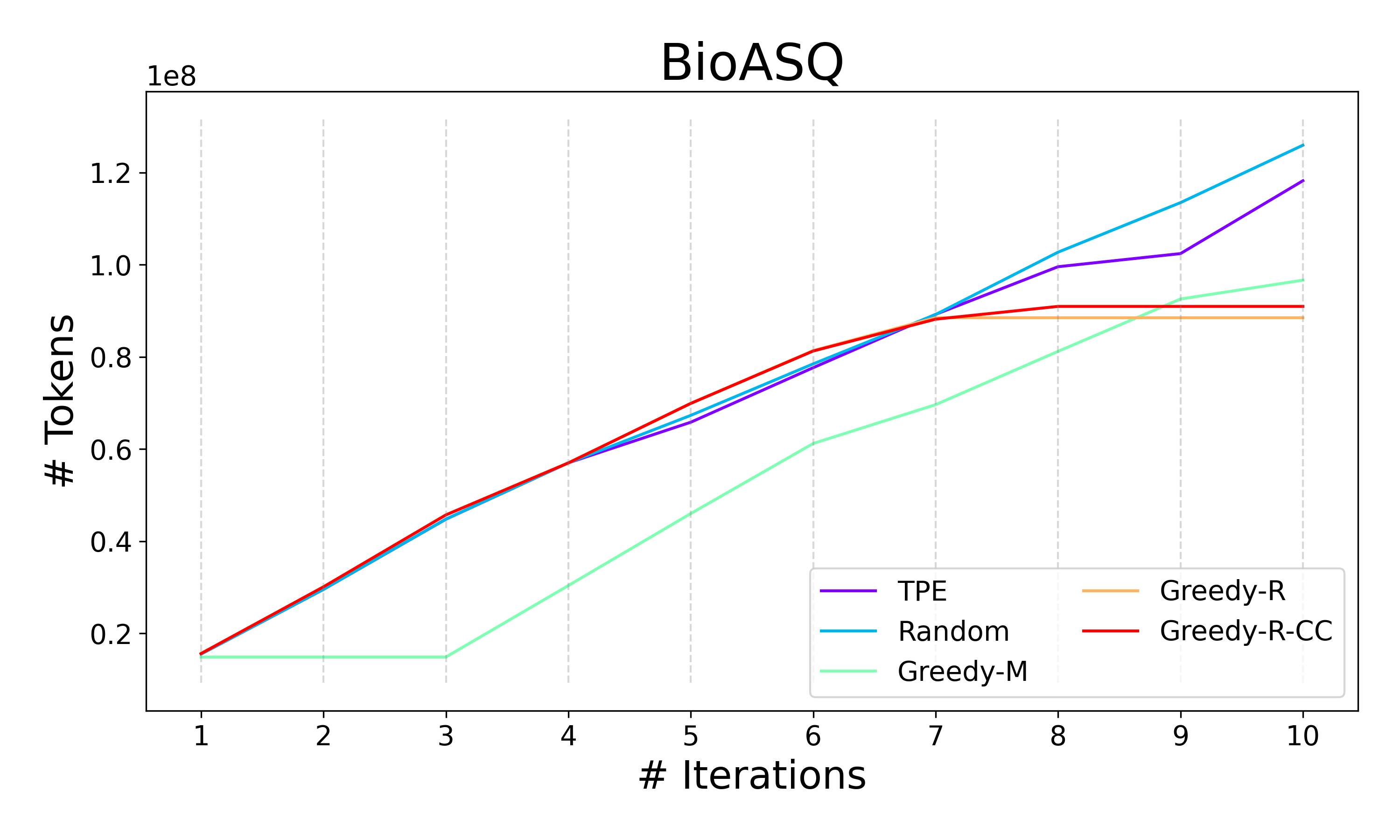} & 
            \includegraphics[trim=1cm 0 0 0, width=0.33\textwidth]{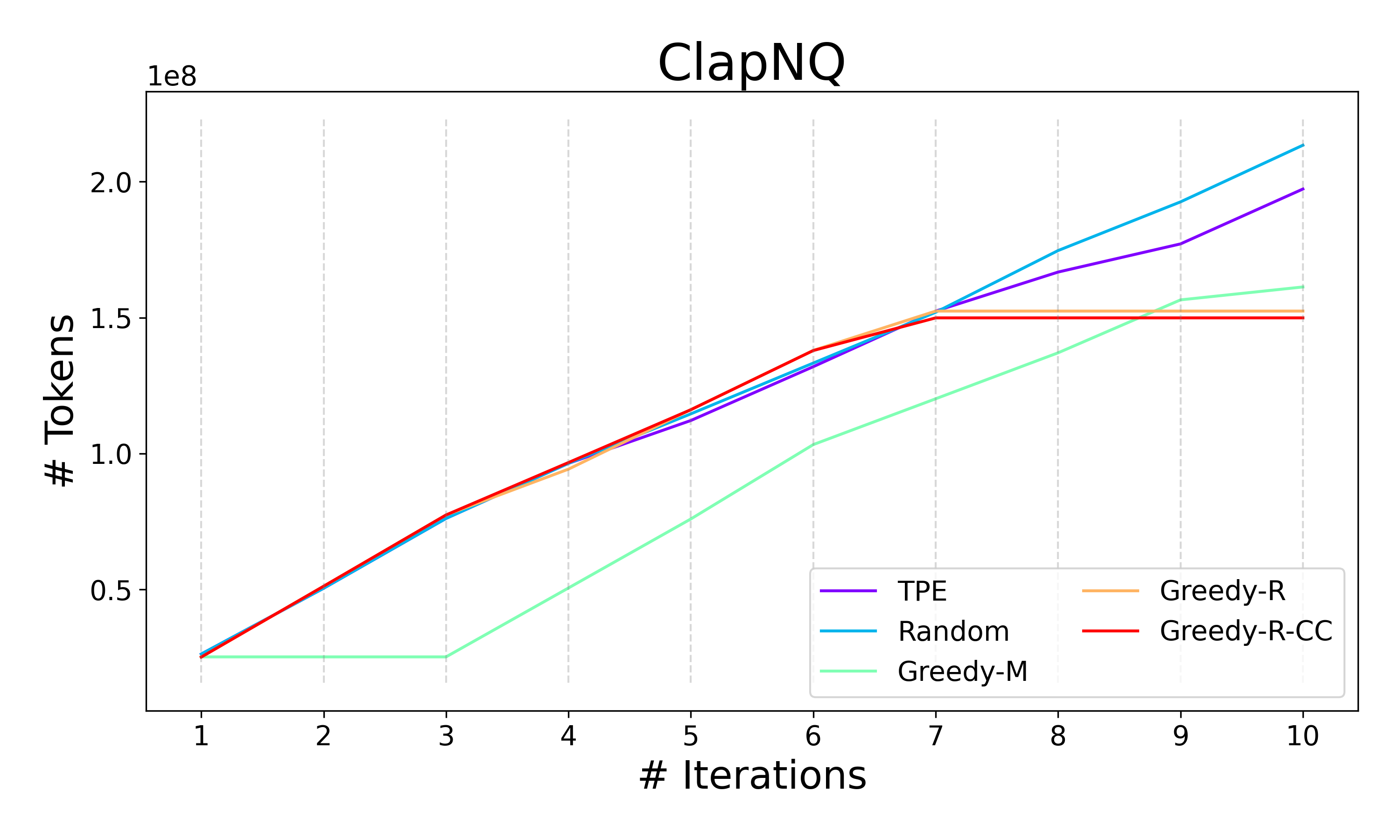} & 
            \includegraphics[trim=1cm 0 0 0, width=0.33\textwidth]{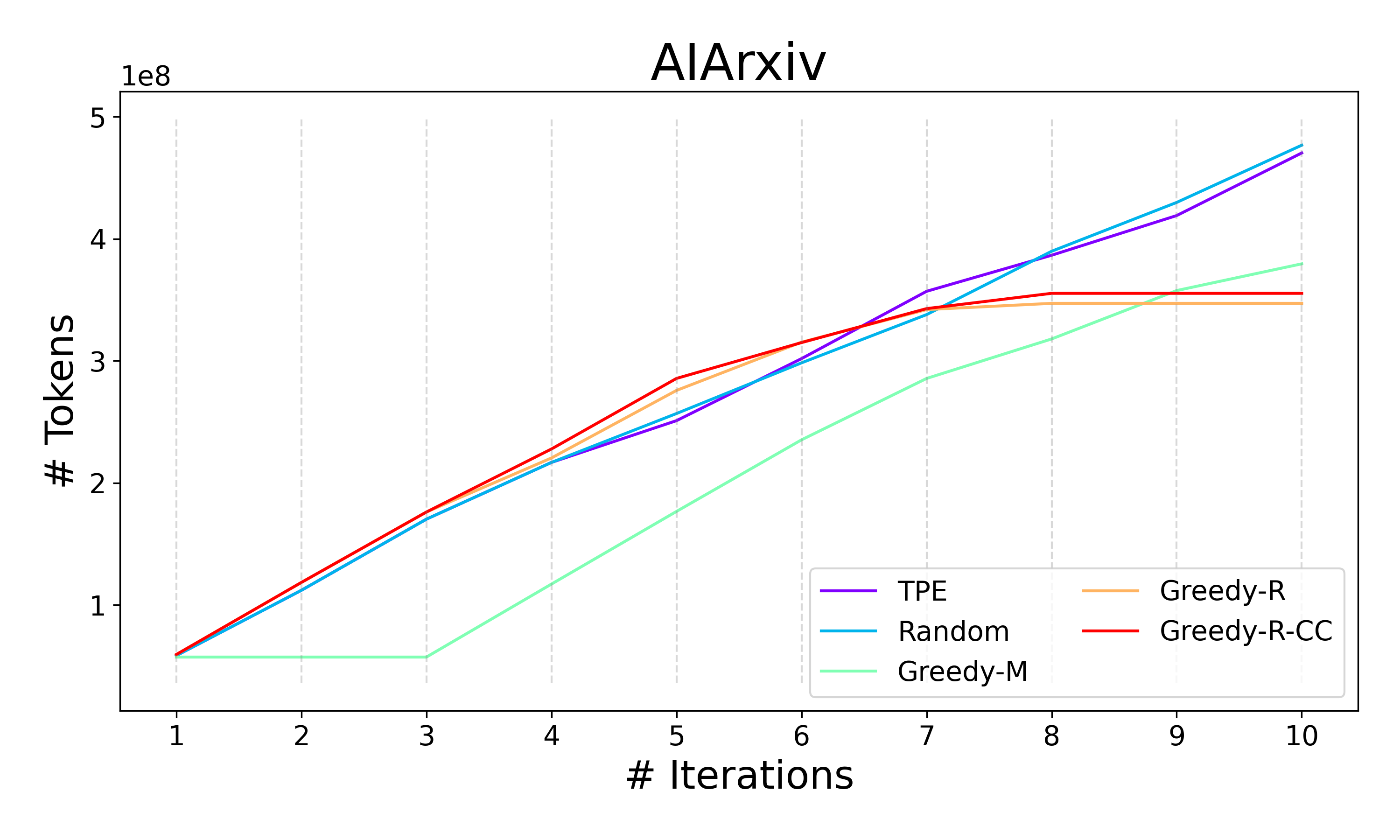} \\
            \includegraphics[trim=1cm 0 0 0, width=0.33\textwidth]{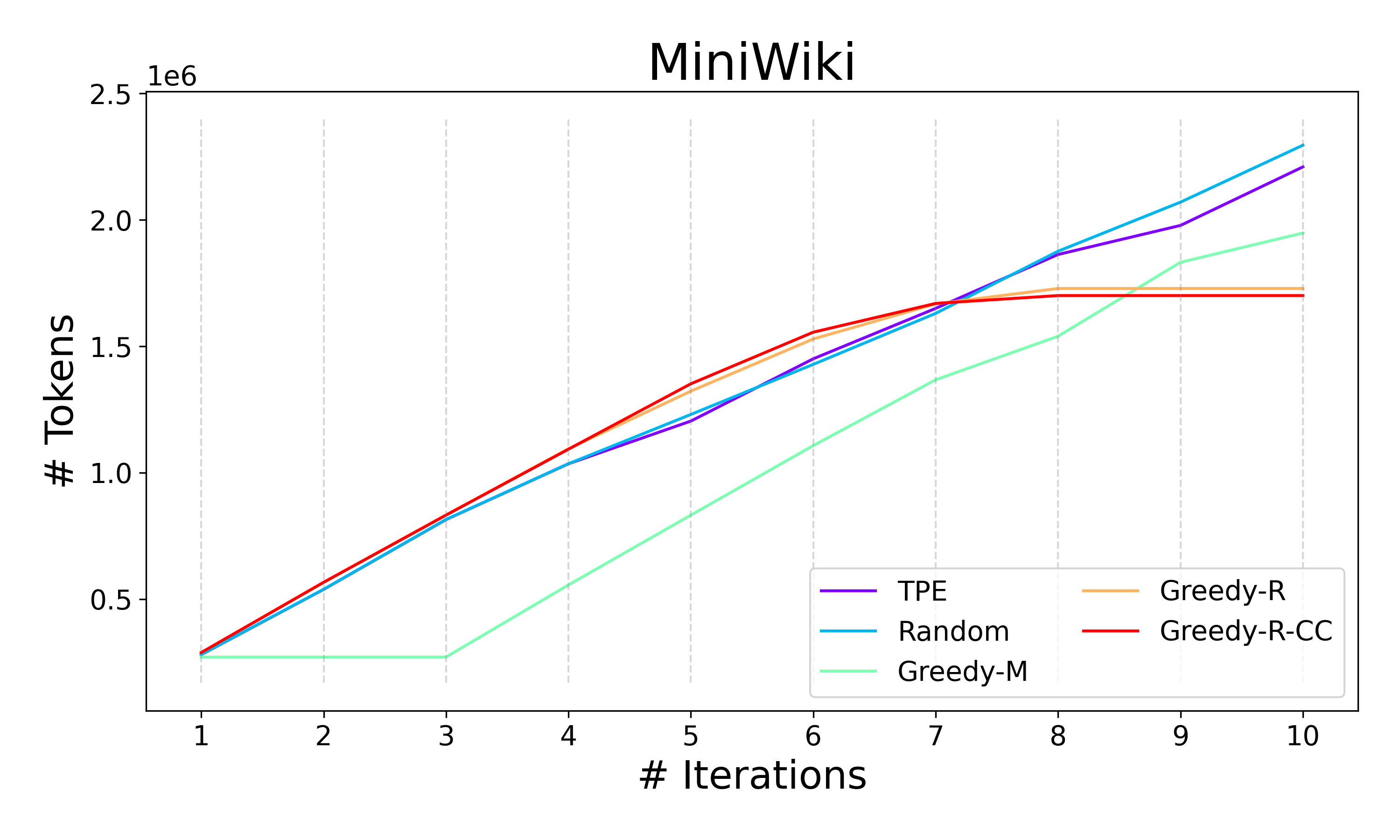} & 
            \includegraphics[trim=1cm 0 0 0, width=0.33\textwidth]{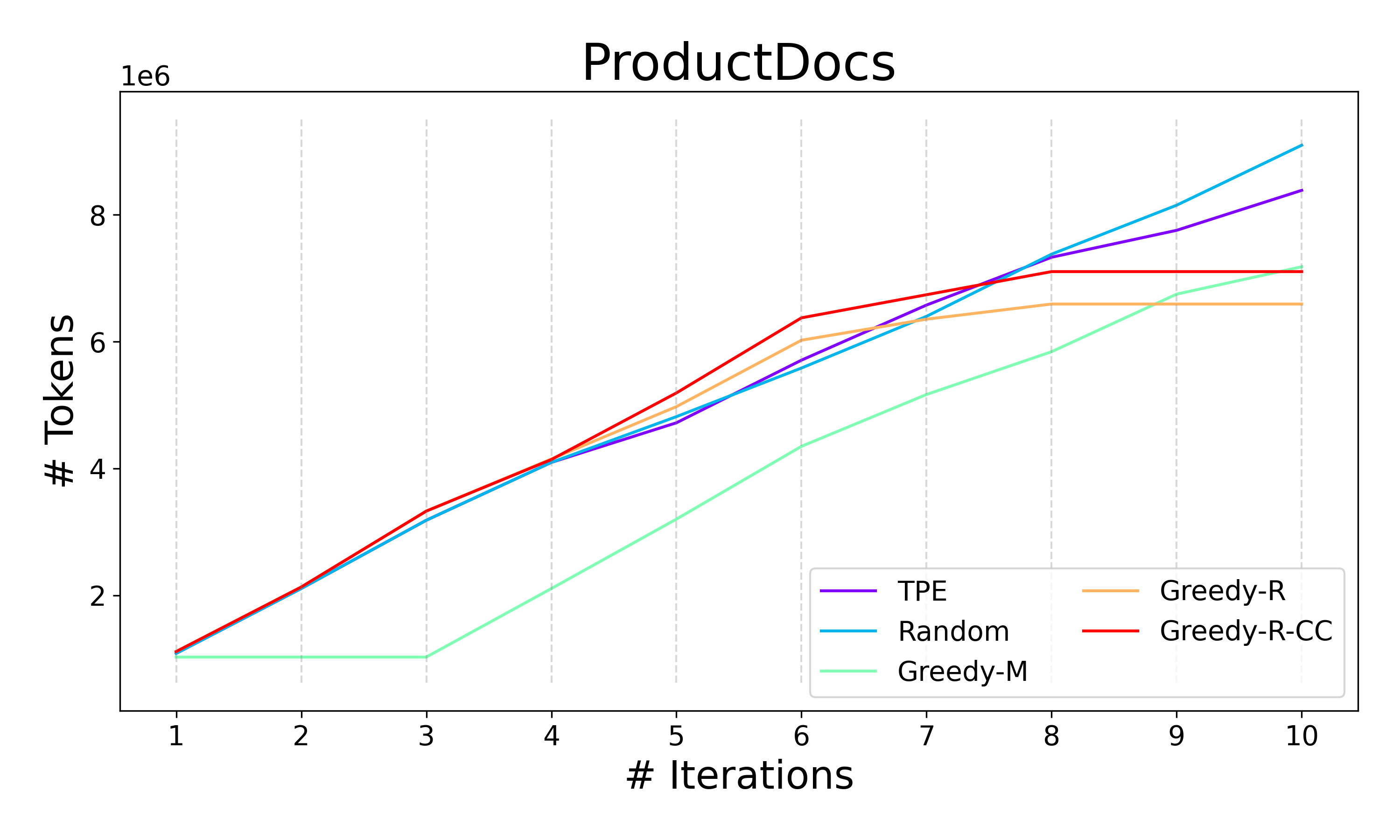} & 
        \end{tabular}
    }
    \vspace{1em}
    
    % Second Subfigure
    \subfigure[Total number of tokens for prompts sent to the generation models.]{
        \centering
        \begin{tabular}{ccc}
            \includegraphics[trim=1cm 0 0 0, width=0.33\textwidth]{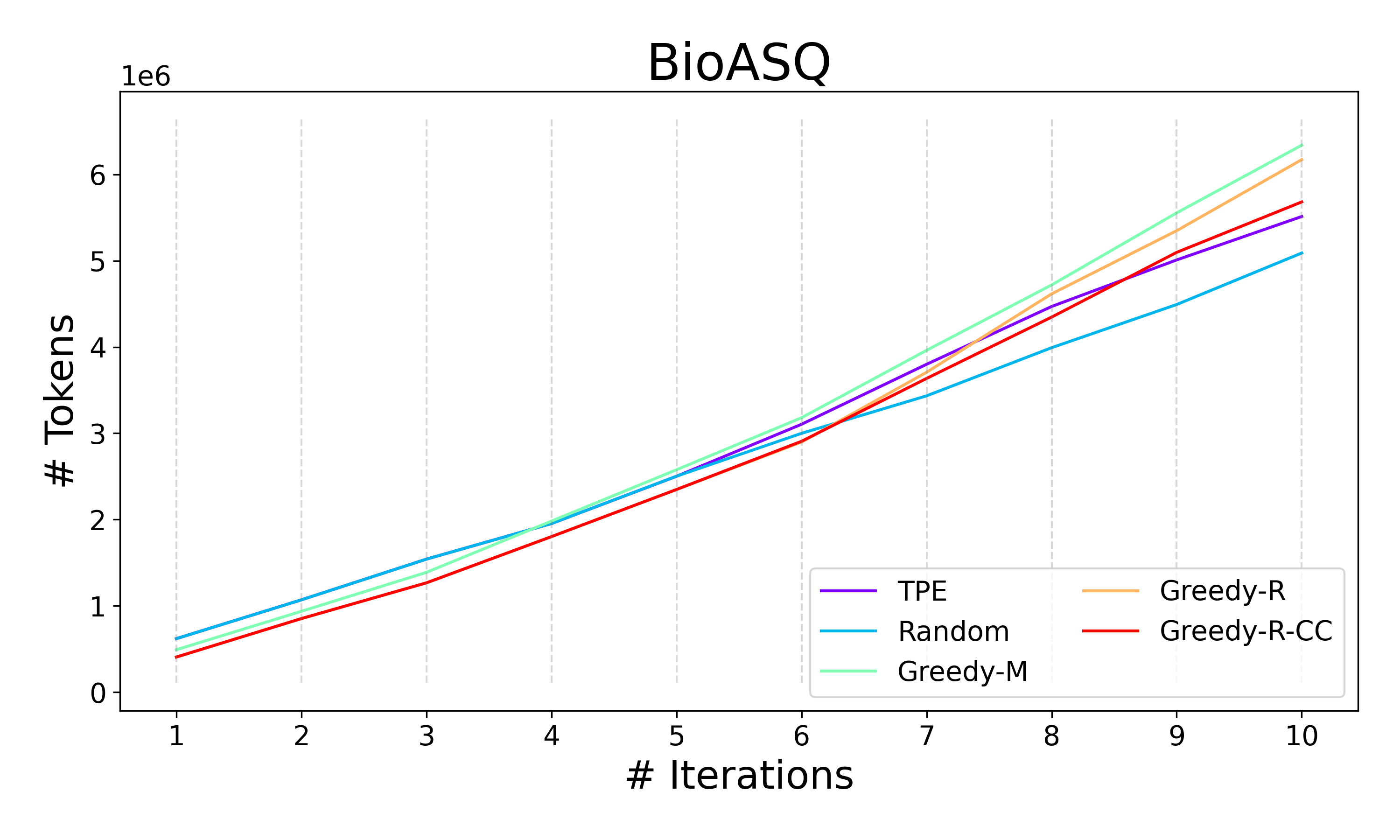} & 
            \includegraphics[trim=1cm 0 0 0, width=0.33\textwidth]{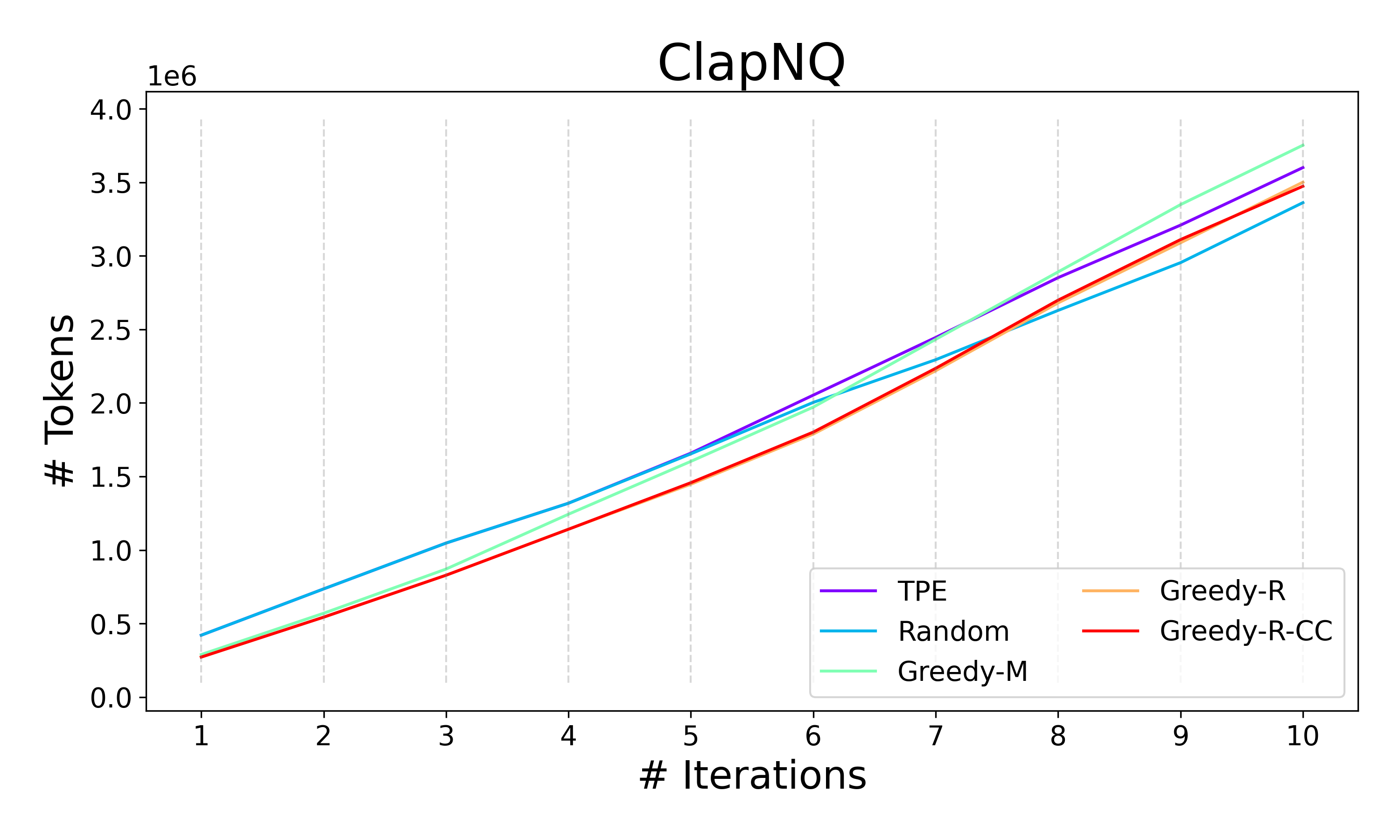} & 
            \includegraphics[trim=1cm 0 0 0, width=0.33\textwidth]{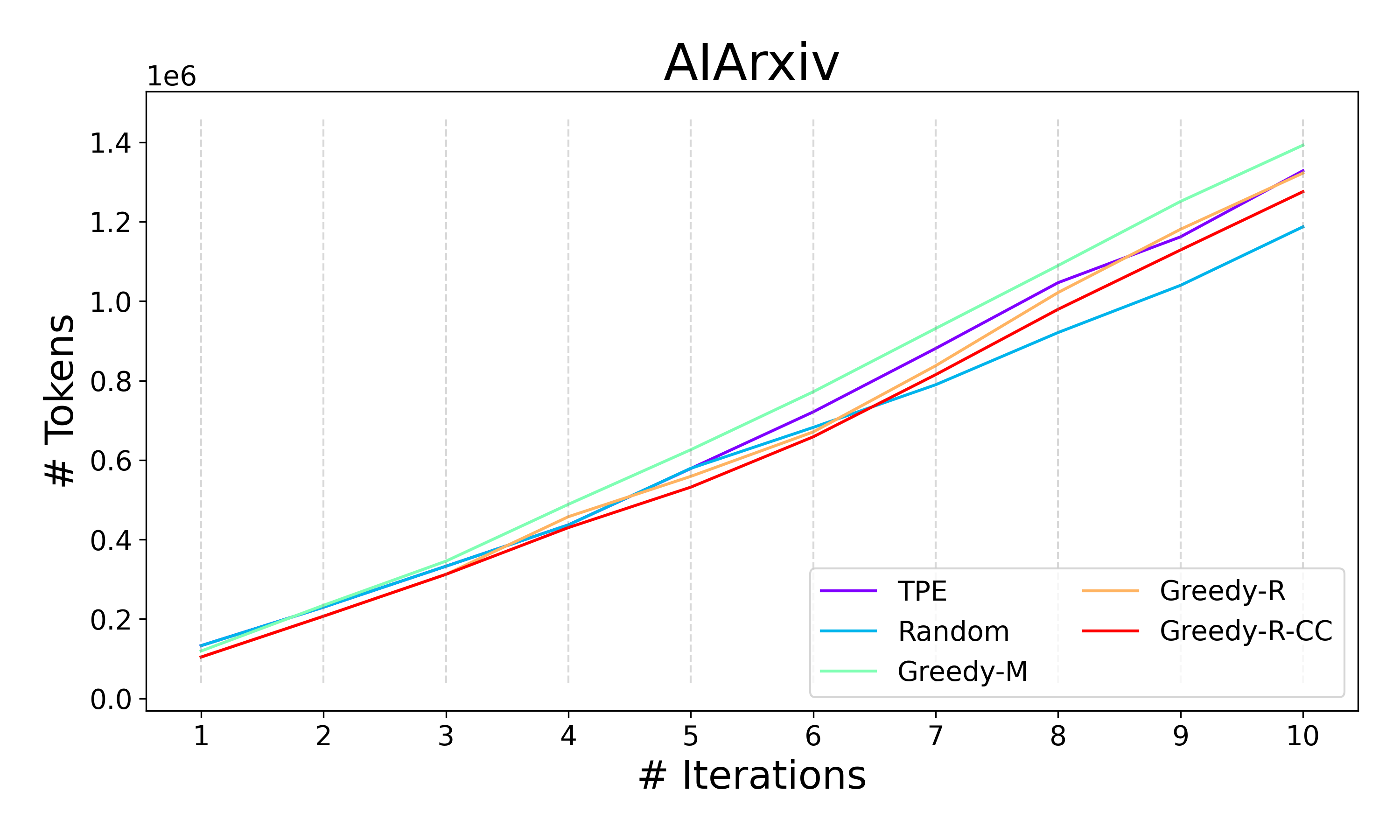} \\
            \includegraphics[trim=1cm 0 0 0, width=0.33\textwidth]{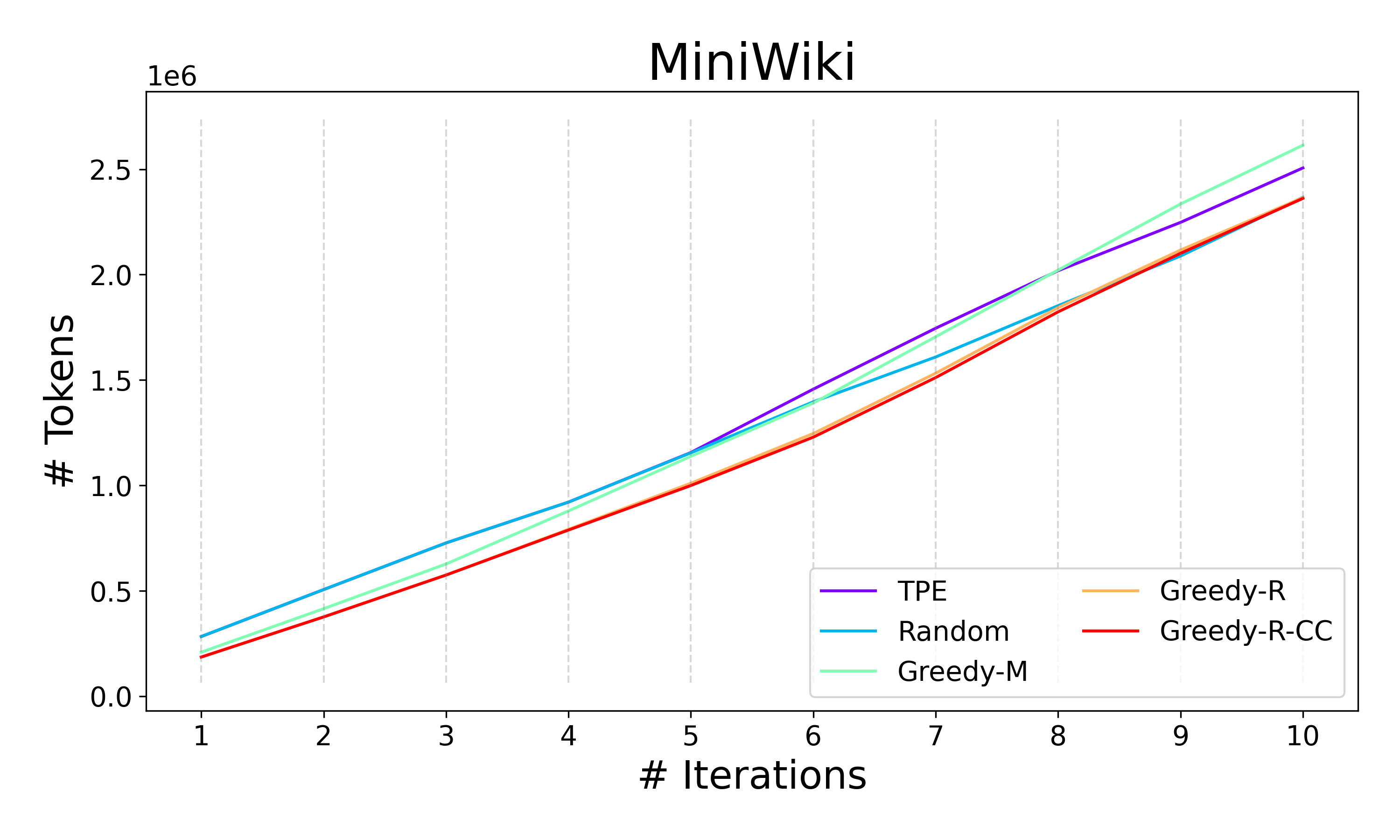} & 
            \includegraphics[trim=1cm 0 0 0, width=0.33\textwidth]{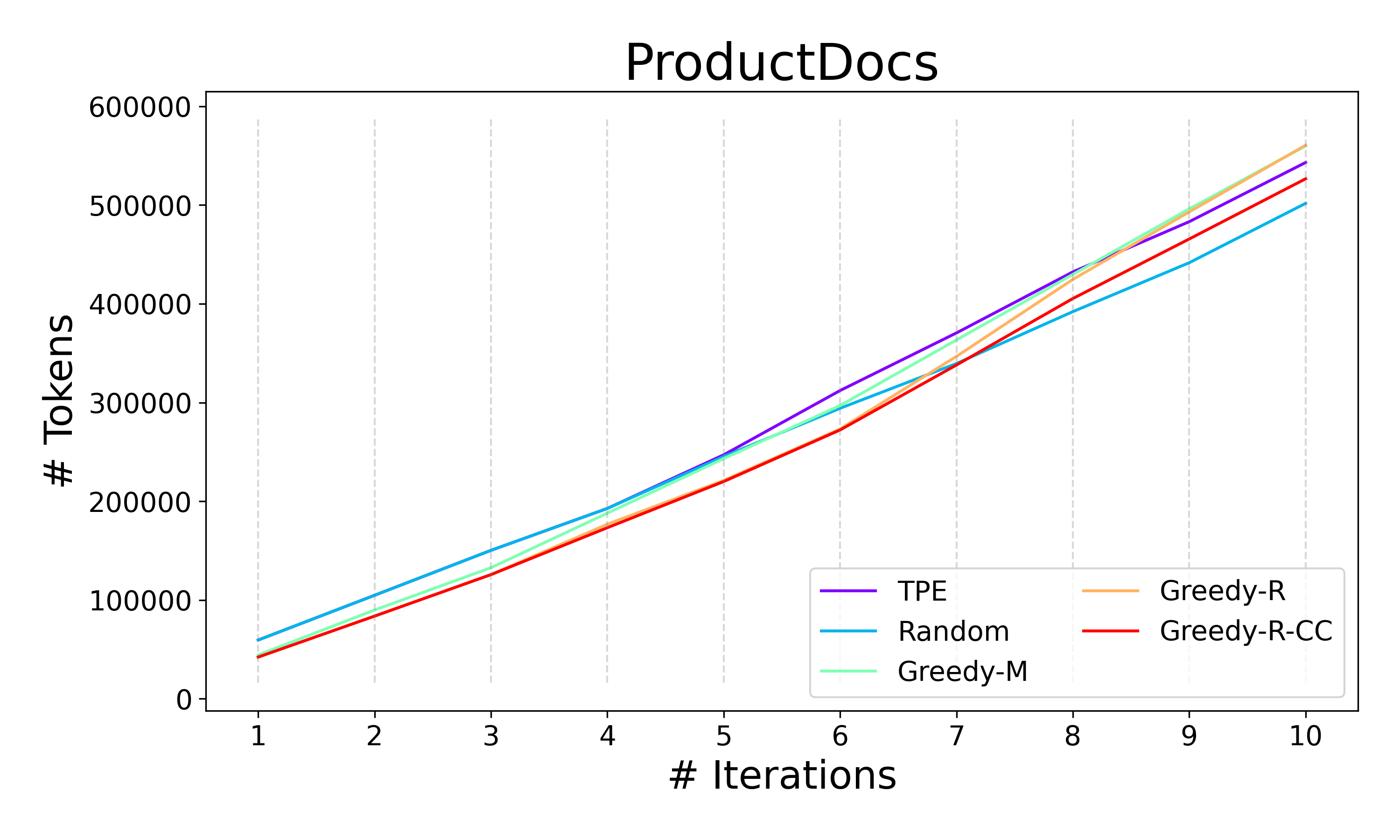} & 
        \end{tabular}
    }
    
    \caption{Cost estimation for each algorithm after each iteration.}
    \label{fig:cost}
\end{figure*}

\figureRef{fig:cost} details accumulated numbers of (a) embedded tokens and (b) number of tokens used in generation part for HPO algorithms overall tested configurations. 
See \sectionRef{ssec:costcons} for a discussion of costs.

\section{Hardware and Costs}
All used embedding and generation models are open source models. An internal in-house infrastructure containing V100 and A100 GPUs was used to run embedding computations and generative inference.
Specifically, embeddings were computed using one V100 GPU, and inference was done on one A100 GPU (i.e. no multi-GPU inference was required).

The evaluation of the \llmaaj metric was done with GPT4o-mini \cite{openai2024gpt4omini} as its backbone LLM. 
That model was used through Microsoft Azure. The overall cost was $\sim500$\$.

% \resizebox{\linewidth}{!}{

\begin{table}
\caption{Results of a likelihood ratio test for the grid-search results of the AIArxiv dataset (\llmaaj{} metric)}
\label{tab:lrt_a}
\resizebox{\linewidth}{!}{
\begin{tabular}{llll}
\toprule
Main effect/Interaction & $\chi^2$ & degrees of freedom & p-value \\
\midrule
Generative model & 401 & 10 & $5.5\times10^{-80}$ \\
Embedding model & 42 & 10 & $9.1\times10^{-6}$ \\
Chunk size & 48 & 8 & $1.2\times10^{-7}$ \\
Chunk overlap & 1.6 & 3 & 0.65 \\
K & 29 & 6 & $5.5\times10^{-5}$ \\
Generative model * Embedding model & 2.2 & 4 & 0.7 \\
Embedding model * Chunk size & 31 & 4 & $3.2\times10^{-6}$ \\
Chunk size * Chunk overlap & 0.28 & 2 & 0.87 \\
Generative model * K & 16 & 4 & $3.5\times10^{-3}$ \\
\bottomrule
\end{tabular}
}
% \end{table}

\medskip
\medskip

% \begin{table}
\caption{Results of a likelihood ratio test for the grid-search results of the BioASQ dataset (\llmaaj{} metric)}
\label{tab:lrt_b}
\resizebox{\linewidth}{!}{
\begin{tabular}{llll}
\toprule
Main effect/Interaction & $\chi^2$ & degrees of freedom & p-value \\
\midrule
Generative model & 5446 & 10 & $\approx{0}$ \\
Embedding model & 251 & 10 & $4.0\times10^{-48}$ \\
Chunk size & 55 & 8 & $5.4\times10^{-9}$ \\
Chunk overlap & 11 & 3 & 0.01 \\
K & 218 & 6 & $2.4\times10^{-44}$ \\
Generative model * Embedding model & 33 & 4 & $1.3\times10^{-6}$ \\
Embedding model * Chunk size & 25 & 4 & $4.6\times10^{-5}$ \\
Chunk size * Chunk overlap & 7.4 & 2 & 0.025 \\
Generative model * K & 150 & 4 & $2.3\times10^{-31}$ \\
\bottomrule
\end{tabular}
}
% \end{table}

\medskip
\medskip

% \begin{table}
\caption{Results of a likelihood ratio test for the grid-search results of the ClapNQ dataset (\llmaaj{} metric)}
\label{tab:lrt_c}
\resizebox{\linewidth}{!}{
\begin{tabular}{llll}
\toprule
Main effect/Interaction & $\chi^2$ & degrees of freedom & p-value \\
\midrule
Generative model & 3103 & 10 & $\approx{0}$ \\
Embedding model & 1883 & 10 & $\approx{0}$ \\
Chunk size & 46 & 8 & $1.9\times10^{-7}$ \\
Chunk overlap & 23 & 3 & $3.3\times10^{-5}$ \\
K & 82 & 6 & $1.4\times10^{-15}$ \\
Generative model * Embedding model & 304 & 4 & $1.7\times10^{-64}$ \\
Embedding model * Chunk size & 9.5 & 4 & 0.049 \\
Chunk size * Chunk overlap & 12 & 2 & $2.5\times10^{-3}$ \\
Generative model * K & 76 & 4 & $1.0\times10^{-15}$ \\
\bottomrule
\end{tabular}
}
% \end{table}

\medskip
\medskip

% \begin{table}
\caption{Results of a likelihood ratio test for the grid-search results of the MiniWiki dataset (\llmaaj{} metric)}
\label{tab:lrt_d}
\resizebox{\linewidth}{!}{
\begin{tabular}{llll}
\toprule
Main effect/Interaction & $\chi^2$ & degrees of freedom & p-value \\
\midrule
Generative model & 12072 & 10 & $\approx{0}$ \\
Embedding model & 140 & 10 & $3.6\times10^{-25}$ \\
Chunk size & 1.6 & 8 & 0.99 \\
Chunk overlap & 0.72 & 3 & 0.87 \\
K & 631 & 6 & $5.4\times10^{-133}$ \\
Generative model * Embedding model & 41 & 4 & $3.4\times10^{-8}$ \\
Embedding model * Chunk size & 0.8 & 4 & 0.94 \\
Chunk size * Chunk overlap & 0.72 & 2 & 0.7 \\
Generative model * K & 626 & 4 & $4.2\times10^{-134}$ \\
\bottomrule
\end{tabular}
}
% \end{table}

\medskip
\medskip

% \begin{table}
\caption{Results of a likelihood ratio test for the grid-search results of the WatsonxQA dataset (\llmaaj{} metric)}
\label{tab:lrt_e}
\resizebox{\linewidth}{!}{
\begin{tabular}{llll}
\toprule
Main effect/Interaction & $\chi^2$ & degrees of freedom & p-value \\
\midrule
Generative model & 757 & 10 & $4.5\times10^{-156}$ \\
Embedding model & 25 & 10 & $5.3\times10^{-3}$ \\
Chunk size & 40 & 8 & $3.5\times10^{-6}$ \\
Chunk overlap & 0.49 & 3 & 0.92 \\
K & 43 & 6 & $1.3\times10^{-7}$ \\
Generative model * Embedding model & 13 & 4 & 0.012 \\
Embedding model * Chunk size & 11 & 4 & 0.026 \\
Chunk size * Chunk overlap & 0.41 & 2 & 0.81 \\
Generative model * K & 7.1 & 4 & 0.13 \\
\bottomrule
\end{tabular}
}
\end{table}

\begin{table}
\caption{Results of a likelihood ratio test for the grid-search results of the AIArxiv dataset (\lexical{} metric)}
\label{tab:lrt_f}
\resizebox{\linewidth}{!}{
\begin{tabular}{llll}
\toprule
Main effect/Interaction & $\chi^2$ & degrees of freedom & p-value \\
\midrule
Generative model & 1918 & 10 & $\approx{0}$ \\
Embedding model & 143 & 10 & $9.5\times10^{-26}$ \\
Chunk size & 68 & 8 & $1.4\times10^{-11}$ \\
Chunk overlap & 1.6 & 3 & 0.66 \\
K & 90 & 6 & $3.7\times10^{-17}$ \\
Generative model * Embedding model & 6.6 & 4 & 0.16 \\
Embedding model * Chunk size & 53 & 4 & $7.1\times10^{-11}$ \\
Chunk size * Chunk overlap & 0.27 & 2 & 0.87 \\
Generative model * K & 26 & 4 & $2.8\times10^{-5}$ \\
\bottomrule
\end{tabular}
}
% \end{table}

\medskip
\medskip

% \begin{table}
\caption{Results of a likelihood ratio test for the grid-search results of the BioASQ dataset (\lexical{} metric)}
\label{tab:lrt_g}
\resizebox{\linewidth}{!}{
\begin{tabular}{llll}
\toprule
Main effect/Interaction & $\chi^2$ & degrees of freedom & p-value \\
\midrule
Generative model & 10487 & 10 & $\approx{0}$ \\
Embedding model & 610 & 10 & $1.3\times10^{-124}$ \\
Chunk size & 126 & 8 & $1.6\times10^{-23}$ \\
Chunk overlap & 3.3 & 3 & 0.35 \\
K & 632 & 6 & $2.6\times10^{-133}$ \\
Generative model * Embedding model & 24 & 4 & $8.2\times10^{-5}$ \\
Embedding model * Chunk size & 58 & 4 & $7.9\times10^{-12}$ \\
Chunk size * Chunk overlap & 3.2 & 2 & 0.2 \\
Generative model * K & 313 & 4 & $2.0\times10^{-66}$ \\
\bottomrule
\end{tabular}
}
% \end{table}

\medskip
\medskip

% \begin{table}
\caption{Results of a likelihood ratio test for the grid-search results of the ClapNQ dataset (\lexical{} metric)}
\label{tab:lrt_h}
\resizebox{\linewidth}{!}{
\begin{tabular}{llll}
\toprule
Main effect/Interaction & $\chi^2$ & degrees of freedom & p-value \\
\midrule
Generative model & 24649 & 10 & $\approx{0}$ \\
Embedding model & 8080 & 10 & $\approx{0}$ \\
Chunk size & 186 & 8 & $5.4\times10^{-36}$ \\
Chunk overlap & 149 & 3 & $4.3\times10^{-32}$ \\
K & 582 & 6 & $1.5\times10^{-122}$ \\
Generative model * Embedding model & 64 & 4 & $4.5\times10^{-13}$ \\
Embedding model * Chunk size & 29 & 4 & $8.3\times10^{-6}$ \\
Chunk size * Chunk overlap & 27 & 2 & $1.1\times10^{-6}$ \\
Generative model * K & 72 & 4 & $8.2\times10^{-15}$ \\
\bottomrule
\end{tabular}
}
% \end{table}

\medskip
\medskip

% \begin{table}
\caption{Results of a likelihood ratio test for the grid-search results of the MiniWiki dataset (\lexical{} metric)}
\label{tab:lrt_i}
\resizebox{\linewidth}{!}{
\begin{tabular}{llll}
\toprule
Main effect/Interaction & $\chi^2$ & degrees of freedom & p-value \\
\midrule
Generative model & 12805 & 10 & $\approx{0}$ \\
Embedding model & 73 & 10 & $1.4\times10^{-11}$ \\
Chunk size & 1.3 & 8 & 1 \\
Chunk overlap & 0.013 & 3 & 1 \\
K & 74 & 6 & $7.3\times10^{-14}$ \\
Generative model * Embedding model & 27 & 4 & $2.5\times10^{-5}$ \\
Embedding model * Chunk size & 1.3 & 4 & 0.87 \\
Chunk size * Chunk overlap & 0.012 & 2 & 0.99 \\
Generative model * K & 25 & 4 & $6.0\times10^{-5}$ \\
\bottomrule
\end{tabular}
}
% \end{table}

\medskip
\medskip

% \begin{table}
\caption{Results of a likelihood ratio test for the grid-search results of the WatsonxQA dataset (\lexical{} metric)}
\label{tab:lrt_j}
\resizebox{\linewidth}{!}{
\begin{tabular}{llll}
\toprule
Main effect/Interaction & $\chi^2$ & degrees of freedom & p-value \\
\midrule
Generative model & 276 & 10 & $2.3\times10^{-53}$ \\
Embedding model & 12 & 10 & 0.28 \\
Chunk size & 70 & 8 & $4.8\times10^{-12}$ \\
Chunk overlap & 2.5 & 3 & 0.48 \\
K & 60 & 6 & $4.1\times10^{-11}$ \\
Generative model * Embedding model & 2.6 & 4 & 0.62 \\
Embedding model * Chunk size & 7.3 & 4 & 0.12 \\
Chunk size * Chunk overlap & 2 & 2 & 0.36 \\
Generative model * K & 8.7 & 4 & 0.068 \\
\bottomrule
\end{tabular}
}
\end{table}

\begin{table*}
\begin{tabular}{llrrrr}
\toprule
 &  & \llmaaj{} & \llmaaj{} ($\Delta$) & \lexical{} & \lexical{} ($\Delta$) \\
\midrule
\multirow[t]{3}{*}{Generative model} & ibm-granite/granite-3.1-8b-instruct & 0.43 & -0.037 & 0.59 & 0.029 \\
 & meta-llama/llama-3-1-8b-instruct & 0.53 & 0.056 & 0.48 & -0.083 \\
 & mistral-nemo-instruct & 0.45 & -0.019 & 0.62 & 0.054 \\
\cline{1-6}
\multirow[t]{3}{*}{Embedding model} & BAAI/bge-large-en-v1.5 & 0.48 & 0.0062 & 0.58 & 0.014 \\
 & ibm/slate-125m-english-rtrvr & 0.47 & 0.0017 & 0.57 & 0.0012 \\
 & intfloat/multilingual-e5-large & 0.46 & -0.0079 & 0.55 & -0.015 \\
\cline{1-6}
\multirow[t]{3}{*}{Chunk size} & 256 & 0.47 & 0.0016 & 0.56 & -0.0024 \\
 & 384 & 0.48 & 0.0092 & 0.57 & 0.0067 \\
 & 512 & 0.46 & -0.011 & 0.56 & -0.0043 \\
\cline{1-6}
\multirow[t]{2}{*}{Chunk overlap} & 0.0 & 0.47 & 0.0024 & 0.57 & 0.0015 \\
 & 0.25 & 0.47 & -0.0024 & 0.56 & -0.0015 \\
\cline{1-6}
\multirow[t]{3}{*}{K} & 3 & 0.46 & -0.0099 & 0.55 & -0.013 \\
 & 5 & 0.47 & 0.0016 & 0.57 & 0.0007 \\
 & 10 & 0.48 & 0.0083 & 0.58 & 0.012 \\
\cline{1-6}
\bottomrule
\end{tabular}
\caption{Marginal means for the grid search results (AIArxiv). Columns denoted by $\Delta$ show the relative difference from the overall dataset mean.}
\label{tab:mmeans_a}
% \end{table}

\medskip
\medskip

% \begin{table}
\begin{tabular}{llrrrr}
\toprule
 &  & \llmaaj{} & \llmaaj{} ($\Delta$) & \lexical{} & \lexical{} ($\Delta$) \\
\midrule
\multirow[t]{3}{*}{Generative model} & ibm-granite/granite-3.1-8b-instruct & 0.45 & -0.042 & 0.61 & 0.035 \\
 & meta-llama/llama-3-1-8b-instruct & 0.53 & 0.036 & 0.52 & -0.053 \\
 & mistral-nemo-instruct & 0.5 & 0.0058 & 0.59 & 0.017 \\
\cline{1-6}
\multirow[t]{3}{*}{Embedding model} & BAAI/bge-large-en-v1.5 & 0.5 & 0.0081 & 0.58 & 0.011 \\
 & ibm/slate-125m-english-rtrvr & 0.49 & -0.0015 & 0.57 & -0.0015 \\
 & intfloat/multilingual-e5-large & 0.49 & -0.0066 & 0.56 & -0.0096 \\
\cline{1-6}
\multirow[t]{3}{*}{Chunk size} & 256 & 0.5 & 0.0015 & 0.57 & -0.0022 \\
 & 384 & 0.49 & -0.0029 & 0.57 & -0.0021 \\
 & 512 & 0.5 & 0.0014 & 0.58 & 0.0042 \\
\cline{1-6}
\multirow[t]{2}{*}{Chunk overlap} & 0.0 & 0.49 & -0.00087 & 0.57 & 0.000067 \\
 & 0.25 & 0.5 & 0.00088 & 0.57 & -0.000071 \\
\cline{1-6}
\multirow[t]{3}{*}{K} & 3 & 0.49 & -0.0041 & 0.56 & -0.0092 \\
 & 5 & 0.5 & -0.00066 & 0.58 & 0.0031 \\
 & 10 & 0.5 & 0.0048 & 0.58 & 0.0061 \\
\cline{1-6}
\bottomrule
\end{tabular}
\caption{Marginal means for the grid search results (BioASQ). Columns denoted by $\Delta$ show the relative difference from the overall dataset mean.}
\label{tab:mmeans_b}
% \end{table*}

\medskip
\medskip

% \begin{table*}
\begin{tabular}{llrrrr}
\toprule
 &  & \llmaaj{} & \llmaaj{} ($\Delta$) & \lexical{} & \lexical{} ($\Delta$) \\
\midrule
\multirow[t]{3}{*}{Generative model} & ibm-granite/granite-3.1-8b-instruct & 0.48 & -0.029 & 0.57 & 0.059 \\
 & meta-llama/llama-3-1-8b-instruct & 0.54 & 0.027 & 0.43 & -0.078 \\
 & mistral-nemo-instruct & 0.51 & 0.0023 & 0.53 & 0.019 \\
\cline{1-6}
\multirow[t]{3}{*}{Embedding model} & BAAI/bge-large-en-v1.5 & 0.52 & 0.0098 & 0.53 & 0.025 \\
 & ibm/slate-125m-english-rtrvr & 0.53 & 0.015 & 0.53 & 0.02 \\
 & intfloat/multilingual-e5-large & 0.49 & -0.024 & 0.46 & -0.045 \\
\cline{1-6}
\multirow[t]{3}{*}{Chunk size} & 256 & 0.51 & -0.0029 & 0.5 & -0.0056 \\
 & 384 & 0.51 & 0.00066 & 0.51 & 0.0022 \\
 & 512 & 0.51 & 0.0023 & 0.51 & 0.0035 \\
\cline{1-6}
\multirow[t]{2}{*}{Chunk overlap} & 0.0 & 0.51 & -0.0015 & 0.5 & -0.0039 \\
 & 0.25 & 0.51 & 0.0015 & 0.51 & 0.0039 \\
\cline{1-6}
\multirow[t]{3}{*}{K} & 3 & 0.51 & 0.0014 & 0.5 & -0.01 \\
 & 5 & 0.51 & -0.00088 & 0.51 & 0.0015 \\
 & 10 & 0.51 & -0.00054 & 0.52 & 0.0089 \\
\cline{1-6}
\bottomrule
\end{tabular}
\caption{Marginal means for the grid search results (ClapNQ). Columns denoted by $\Delta$ show the relative difference from the overall dataset mean.}
\label{tab:mmeans_c}
\end{table*}

% \medskip

\begin{table*}
\begin{tabular}{llrrrr}
\toprule
 &  & \llmaaj{} & \llmaaj{} ($\Delta$) & \lexical{} & \lexical{} ($\Delta$) \\
\midrule
\multirow[t]{3}{*}{Generative model} & ibm-granite/granite-3.1-8b-instruct & 0.36 & -0.081 & 0.84 & 0.084 \\
 & meta-llama/llama-3-1-8b-instruct & 0.49 & 0.05 & 0.62 & -0.13 \\
 & mistral-nemo-instruct & 0.47 & 0.031 & 0.8 & 0.048 \\
\cline{1-6}
\multirow[t]{3}{*}{Embedding model} & BAAI/bge-large-en-v1.5 & 0.44 & 0.00058 & 0.75 & -0.0012 \\
 & ibm/slate-125m-english-rtrvr & 0.43 & -0.0066 & 0.75 & -0.0061 \\
 & intfloat/multilingual-e5-large & 0.44 & 0.0061 & 0.76 & 0.0071 \\
\cline{1-6}
\multirow[t]{3}{*}{Chunk size} & 256 & 0.44 & -0.000017 & 0.75 & -0.00016 \\
 & 384 & 0.44 & -0.000067 & 0.75 & 0.000091 \\
 & 512 & 0.44 & 0.00015 & 0.75 & 0.000012 \\
\cline{1-6}
\multirow[t]{2}{*}{Chunk overlap} & 0.0 & 0.44 & 0.000024 & 0.75 & -0.000054 \\
 & 0.25 & 0.44 & 0.000018 & 0.75 & 0.000014 \\
\cline{1-6}
\multirow[t]{3}{*}{K} & 3 & 0.44 & 0.00085 & 0.74 & -0.008 \\
 & 5 & 0.44 & -0.0016 & 0.75 & 0.003 \\
 & 10 & 0.44 & 0.00086 & 0.76 & 0.005 \\
\cline{1-6}
\bottomrule
\end{tabular}
\caption{Marginal means for the grid search results (MiniWiki). Columns denoted by $\Delta$ show the relative difference from the overall dataset mean.}
\label{tab:mmeans_d}
% \end{table*}

\medskip
\medskip

% \begin{table*}
\begin{tabular}{llrrrr}
\toprule
 &  & \llmaaj{} & \llmaaj{} ($\Delta$) & \lexical{} & \lexical{} ($\Delta$) \\
\midrule
\multirow[t]{3}{*}{Generative model} & ibm-granite/granite-3.1-8b-instruct & 0.59 & -0.041 & 0.83 & 0.034 \\
 & meta-llama/llama-3-1-8b-instruct & 0.7 & 0.074 & 0.78 & -0.018 \\
 & mistral-nemo-instruct & 0.6 & -0.032 & 0.78 & -0.016 \\
\cline{1-6}
\multirow[t]{3}{*}{Embedding model} & BAAI/bge-large-en-v1.5 & 0.63 & 0.00097 & 0.8 & -0.00023 \\
 & ibm/slate-125m-english-rtrvr & 0.63 & -0.0028 & 0.8 & -0.0025 \\
 & intfloat/multilingual-e5-large & 0.63 & 0.0019 & 0.8 & 0.0028 \\
\cline{1-6}
\multirow[t]{3}{*}{Chunk size} & 256 & 0.62 & -0.0089 & 0.79 & -0.01 \\
 & 384 & 0.63 & -0.005 & 0.79 & -0.006 \\
 & 512 & 0.64 & 0.014 & 0.82 & 0.016 \\
\cline{1-6}
\multirow[t]{2}{*}{Chunk overlap} & 0.0 & 0.63 & -0.00052 & 0.8 & -0.00098 \\
 & 0.25 & 0.63 & 0.00053 & 0.8 & 0.00098 \\
\cline{1-6}
\multirow[t]{3}{*}{K} & 3 & 0.65 & 0.016 & 0.79 & -0.014 \\
 & 5 & 0.62 & -0.0055 & 0.8 & 0.003 \\
 & 10 & 0.62 & -0.01 & 0.81 & 0.011 \\
\cline{1-6}
\bottomrule
\end{tabular}
\caption{Marginal means for the grid search results (WatsonxQA). Columns denoted by $\Delta$ show the relative difference from the overall dataset mean.}
\label{tab:mmeans_e}
\end{table*}

\section{Generation Prompt Details}
\label{sec:app_generation_details}
%Greedy decoding was used throughout all experiments. 
%The prompts were fixed to RAG prompts tailored to each model: 
The RAG prompts used by each model are shown in
\figureRef{fig:granite_prompt} for \granite, \figureRef{fig:lamma_prompt} for \lamma and \figureRef{fig:mistral_prompt} for \mistral. 
In each prompt the \emph{\{question\}} placeholder indicates where the user question was placed, and \emph{\{retrieved documents\}} the location of the retrieved chunks. For \granite, each retrieved chunk was prefixed with `\emph{[Document]}' and suffixed by `\emph{[End]}'. Similarly, for \lamma each retrieved chunk was prefixed with `\emph{[document]:}'.

\section{Use Of AI Assistants}
AI Assistants were only used in writing for minor edits and rephrases. They were also used to aid in obtaining the correct LateX syntax for the various figures.

\begin{figure*}[htbp]
  \centering
  \begin{tcolorbox}[halign=flush left, width=0.9\linewidth]
  \small
\texttt{<|system|>}\\

You are Granite Chat, an AI language model developed by IBM. 
You are a cautious assistant. You carefully follow instructions.  
You are helpful and harmless and you follow ethical guidelines and 
promote positive behavior.\\

\texttt{<|user|>}\\

You are a AI language model designed to function as a specialized  Retrieval Augmented Generation (RAG) assistant. \\ 
When generating responses, prioritize correctness, i.e., ensure that 
your response is grounded in context and user query. \\
Always make sure that your response is relevant to the question.\\

Answer Length: detailed\\

{[}Document] \\
\{retrieved documents\} \\
{[}End]\\

\{question\}\\

\texttt{<|assistant|>}
\end{tcolorbox}
    \caption{The prompt used for \granite.}
    \label{fig:granite_prompt}
\end{figure*}

\begin{figure*}[htbp]
  \centering
  \begin{tcolorbox}[halign=flush left, width=1\linewidth]
  \small
\texttt{<|begin\_of\_text|><|start\_header\_id|>system<|end\_header\_id|>}\\
You are a helpful, respectful and honest assistant.\\
Always answer as helpfully
as possible, while being safe. \\
Your answers should not include any harmful,
unethical, racist, sexist, toxic, dangerous, or illegal content. \\
Please ensure
that your responses are socially unbiased and positive in nature.\\
If a question does not make any sense, or is not factually coherent, explain why
instead of answering something not correct.\\
If you don't know the answer to a
question, please don't share false information.\\
\texttt{<|eot\_id|><|start\_header\_id|>user<|end\_header\_id|>}\\

{[}document]: \{retrieved documents\}\\

{[}conversation]: \{question\}. Answer with no more than 150 words.  If you cannot base
your answer on the given document, please state that you do not have an answer.<|eot\_id|>\\
\texttt{<|start\_header\_id|>assistant<|end\_header\_id|>}
    \end{tcolorbox}
    \caption{The prompt used for \lamma.}
    \label{fig:lamma_prompt}
\end{figure*}

\begin{figure*}[htbp]
  \centering
  \begin{tcolorbox}[halign=flush left, width=1\linewidth]
      {\small
\texttt{<s>{[}INST] <<SYS>>}\\

You are a helpful, respectful and honest assistant.\\ 
Always answer as helpfully as possible, while being safe.\\ 
Your answers should not include any harmful, unethical, racist, sexist, 
toxic, dangerous, or illegal content. \\
Please ensure that your responses are socially unbiased and positive in nature.\\
If a question does not make any sense, or is not factually coherent, explain why
instead of answering something not correct. \\ 
If you don't know the answer to a question, please don't share false information.\\
\texttt{<</SYS>>}\\
 Generate the next agent response by answering the question. You are provided several
documents with titles. \\
If the answer comes from different documents please mention all possibilities
and use the titles of documents  to separate between topics or domains. \\If you cannot base your answer
on the given documents, please state that you do not have an answer.\\
 \{retrieved documents\}\\
 \{question\} {[/}INST]
    }
    \end{tcolorbox}
    \caption{The prompt used for \mistral.}
    \label{fig:mistral_prompt}
\end{figure*}

% From here, Benjamin

\begin{figure*}[htbp]

% First horizontal bar
\noindent\rule{\textwidth}{0.4pt}

% Four lines of formatted text
\textbf{Question}: What are the natural language processing tasks supported in the Product library? \par
\textbf{Gold answer}: The Product library supports the following natural language processing tasks: language detection, syntax analysis, noun phrase extraction, keyword extraction and ranking, entity extraction, sentiment classification, and tone classification. \par
\textbf{Gold context-id}: EN-001 \par
\textbf{Gold passage}: The following natural language processing tasks are supported as blocks or workflows in the Product library:
\\

*  [Language detection]\\
*  [Noun phrase extraction]\\
*  [Keyword extraction and ranking]\\
*  [Entity extraction]\\
*  [Sentiment classification]\\
*  [Tone classification] \par

% Second horizontal bar
\noindent\rule{\textwidth}{0.4pt}
\textbf{Question}: 	What happens to unsaved prompt text within Product, and how long does it persist on the webpage before being deleted? \par
\textbf{Gold answer}: The prompt text remains unsaved unless the user decides to save their progress. While unsaved, the prompt text persists on the webpage until a page refresh occurs, upon which the text is automatically deleted. \par
\textbf{Gold context-id}: EN-101 \par
\textbf{Gold passage}: Privacy of text in Product during a session \\

Text that you submit by clicking Generate from the prompt editor in Product is reformatted as tokens, and then submitted to the foundation model you choose. The submitted message is encrypted in transit.\\
\\
Your prompt text is not saved unless you choose to save your work.\\
\\
Unsaved prompt text is kept in the web page until the page is refreshed, at which time the prompt text is deleted. \par

% Third horizontal bar
\noindent\rule{\textwidth}{0.4pt}
    \caption{Examples of benchmark entries in \watsonxdataset.}
    \label{fig:product_sample}
\end{figure*}

\begin{figure*}[htbp]
  \centering
  \begin{tcolorbox}[halign=flush left, width=1\linewidth]
    {\small
    Given the next \texttt{[document]}, create a \texttt{[question]} and \texttt{[answer]} pair that are grounded in the main point of the document, don't add any additional information that is not in the document. The \texttt{[question]} is by an information-seeking User and the \texttt{[answer]} is provided by a helping AI Agent.\\

    \texttt{[document]}: [A \watsonxdataset document example] \\

    \# Response:\\
    \texttt{[question]}: What is a token limit?\\
    \texttt{[answer]}: Every model has an upper limit to the number of tokens in the input prompt plus the number of tokens in the generated output from the model (sometimes called context window length, context window, context length, or maximum sequence length.)\\
    ...\\
    \texttt{[document]}: 
    }
  \end{tcolorbox}
  \caption{The prompt used for synthetic data generation. The prompt consists of instruction followed by $K=3$ examples of document and generated QA.}
  \label{fig:synthetic_prompt}
\end{figure*}

\end{document}